\documentclass{article}
\usepackage{iclr2027_conference,times}

\usepackage{amsmath,amsfonts,bm}

\def\eqref#1{equation~\ref{#1}}

\def\1{\bm{1}}

\DeclareMathAlphabet{\mathsfit}{\encodingdefault}{\sfdefault}{m}{sl}
\SetMathAlphabet{\mathsfit}{bold}{\encodingdefault}{\sfdefault}{bx}{n}

\usepackage{hyperref}
\usepackage{xurl}
\usepackage{url}
\usepackage{booktabs}
\usepackage{array}
\usepackage{graphicx}
\usepackage{amsmath,amssymb}
\usepackage{xcolor}
\usepackage{tikz}
\usetikzlibrary{arrows.meta,positioning,fit,backgrounds}
\usepackage{enumitem}
\usepackage[most]{tcolorbox}
\usepackage{etoolbox}

\makeatletter
\patchcmd{\@maketitle}{\rule{\z@}{24pt}\@author}{\rule{\z@}{19pt}\@author}{}%
  {\PackageError{preprint}{Could not adjust the title-to-author gap}{}}
\patchcmd{\abstract}{\vspace{0.5ex}}{\vspace{-0.7ex}}{}%
  {\PackageError{preprint}{Could not adjust the Abstract heading gap}{}}
\patchcmd{\endabstract}{\vskip 1ex}{\vskip -1pt}{}%
  {\PackageError{preprint}{Could not adjust the gap after Abstract}{}}
\makeatother

\usepackage{etoc}
\usepackage{minitoc}
\etocsettocstyle{\section*{Appendix}}{}

\setlist[itemize]{leftmargin=3.5mm, labelsep=0.5em, topsep=0.4ex, itemsep=0.5ex, parsep=0pt}
\setlist[enumerate]{leftmargin=1.4em, labelsep=0.5em, topsep=0.4ex, itemsep=0.5ex, parsep=0pt}

\definecolor{linkcol}{HTML}{bd4040}   %
\definecolor{citecol}{HTML}{3F7A57}   %
\definecolor{urlcol}{HTML}{6A4E8E}    %
\hypersetup{
  colorlinks=true,
  linkcolor=linkcol,
  citecolor=citecol,
  urlcolor=urlcol,
}

\newif\ifswmorevision\swmorevisionfalse
\definecolor{swmoblue}{HTML}{0055FF}

\newcommand{\sname}{LoC}

\newcommand{\llname}{Loud and Clear}

\title{\llname: Making Visual Evidence\\ Count in Multimodal Document RAG}

\author{Eunchan Jo\thanks{Equal contribution.\quad
Code: \url{https://github.com/Chipmunk-g4/Loud-and-Clear}.}\qquad
Seounghyun Jang\footnotemark[2]\qquad Sangwoo Mo\\
POSTECH\\
\texttt{\{eunchan9029, shyun1751, sangwoo.mo\}@postech.ac.kr}
}

\iclrfinalcopy %

\begin{document}
\setcounter{footnote}{1} %
\vspace*{\dimexpr-5pt-\baselineskip\relax} %
\maketitle
\lhead{arXiv Preprint} %
\vspace{-5pt} %

\etocdepthtag.toc{mtchapter}
\etocsettagdepth{mtchapter}{subsection}
\etocsettagdepth{mtappendix}{none}
\faketableofcontents

\begin{abstract}

Multimodal document retrieval-augmented generation (RAG) has focused largely on retrieval, yet its answers depend on how well a vision-language model reads the retrieved pages.
We find that evidence in retrieved pages remains underused even when the correct pages are retrieved.
On SciEGQA oracle pages, five prior methods give the generator 0.7K to 2.7K visual tokens per query on average under a common image processor, yet raising the page-only budget from 1K to 32K improves accuracy by 15.22 points.
At a fixed 16K budget, adding ground-truth crops improves accuracy by 3.75 points over page-only input.
We study what the generator needs from retrieved pages through two requirements: \textbf{Loud}, making full use of visual token capacity, and \textbf{Clear}, refining the page so that its evidence becomes prominent.
For Loud, increasing either the visual token budget or visual fidelity, the image resolution those tokens preserve, improves accuracy with the other held fixed, and full-page scaling outperforms repetition and tiling at matched budgets.
For Clear, suppressing surrounding content improves accuracy without enlarging selected regions, and adding OCR of these regions to page images outperforms either input alone.
We implement both as \textbf{\llname{} (\sname{})}, which combines higher-resolution pages with OCR from query-relevant regions without ground-truth annotations or additional training.
\sname{} outperforms eight prior methods on SciEGQA, ViDoRe-v3, and MMLongBench-Doc, improving over the strongest, ColQwen3, by 8.88, 3.09, and 4.22 points with the same retriever and generator.
It also outperforms its page-only variant under the same visual token budget.
These results locate a bottleneck beyond retrieval: how the retrieved pages are given to the generator.

\end{abstract}

\section{Introduction}
\label{sec:introduction}
\vspace{-5pt} %

Multimodal document RAG extends retrieval-augmented generation (RAG)~\citep{lewis2020retrieval} to visually rich documents~\citep{ICLR2025_3640a199,Tanaka_2025_CVPR}, retrieving relevant pages and passing them to a vision-language model (VLM) as evidence conveyed through text, layout, tables, charts, figures, and typography~\citep{Mathew_2021_WACV,Mathew_2022_WACV}.
Visual retrievers such as ColPali~\citep{ICLR2025_99e9e141} operate directly on page images, preserving layout and other visual signals without intermediate text extraction.
A growing line of end-to-end systems has extended visual retrieval and generation to long, multi-page, and multi-document settings~\citep{Cho_2025_ICCV,ICLR2025_0d40c7b4,suri-etal-2025-visdom,wang-etal-2025-vidorag}.
Much of this work has focused on improving retrieval.
How retrieved pages are then given to the generator remains comparatively underexplored, and many systems pass them at a default resolution without refining their content.

\begin{figure}[t]
    \centering
    \includegraphics[width=\linewidth]{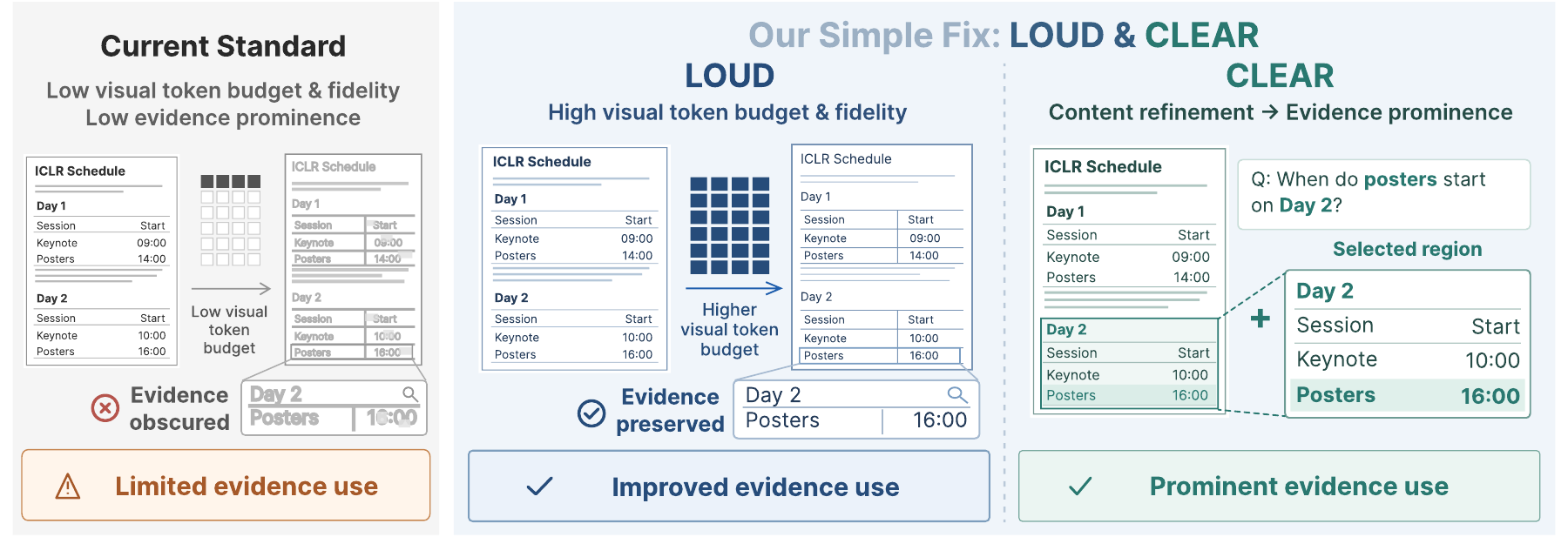}
    \vspace{-17pt}
    \caption{\textbf{Loud and Clear let the generator use evidence that a correctly retrieved page already contains.} For the Day 2 poster question, the current standard's low visual token budget obscures relevant rows. Loud preserves them with a larger budget and higher visual fidelity, while Clear makes the selected region prominent alongside the page. The retrieved page is unchanged throughout.}
    \label{fig:retrieval_usability}
    \vspace{-17pt}
\end{figure}

However, even when retrieval succeeds, the generator underuses the evidence in the retrieved pages.
On SciEGQA~\citep{yu2026sciegqadatasetscientificevidencegrounded} oracle pages, five prior methods give the generator 0.7K to 2.7K visual tokens per query on average when adapted to a common image processor.
Yet with the same pages and generator, scaling the page-only budget from 1K to 32K improves accuracy by 15.22 percentage points.
How the page content is refined also matters: at a fixed 16K budget, pairing pages with ground-truth crops adds 3.75 points over pages alone.
These gains with correct pages and a fixed generator reveal a bottleneck in how retrieved pages are given to the generator.

The systems that do change how retrieved pages are given to the generator leave open which changes help it use the same retrieved pages.
They aggregate pages~\citep{ICLR2025_3640a199}, fuse textual and visual evidence~\citep{suri-etal-2025-visdom}, or replace full pages with retrieved regions~\citep{li2026regionrag}, each within its own pipeline with its own retriever and generator.
The closest system, RegionRAG~\citep{li2026regionrag}, replaces pages with regions, whereas our controlled comparisons show that keeping the page in the input alongside its regions is more reliable across benchmarks.

To address this gap, we organize our diagnosis into two generator input requirements, Loud and Clear, and implement them as the \textbf{\llname{} (\sname{})} framework (Figure~\ref{fig:retrieval_usability}).
\textbf{Loud} fully exploits the generator's \emph{visual token capacity}, its ability to read pages encoded as visual tokens.
It requires a sufficient visual token budget and high \emph{visual fidelity}, how much image resolution survives resizing into visual tokens.
\textbf{Clear} applies \emph{content refinement}, such as whiteout, crop, or OCR, to make each page's evidence prominent.
Holding the retrieved pages and generator fixed, we compare alternative implementations of each requirement on SciEGQA~\citep{yu2026sciegqadatasetscientificevidencegrounded} and ViDoRe-v3~\citep{loison-etal-2026-vidore}, using ground-truth region localization only for oracle diagnosis, not for LoC.

Our controlled comparisons select full-page scaling for Loud and page images with OCR of the selected regions for Clear.
Full-page scaling raises both the visual token budget and visual fidelity, and each improves accuracy with the other held fixed.
Spending the same budget on repeated pages, repeated blank pages, or spatial tiling does not match full-page scaling, so token count alone does not explain the gain.
For Clear, adding OCR of the selected regions to page images outperforms either input alone under the same visual token budget.
Whiting out content outside the selected regions also improves accuracy, whereas extracting or regrouping the regions gives no consistent advantage, which is consistent with keeping the retrieved page in the generator's input.

\sname{} combines full-page scaling with OCR from regions scored using the retriever's token embeddings, without ground-truth annotations or additional training.
\sname{} outperforms eight prior methods, improving over the ColQwen3~\citep{huang2025beyond} baseline with the same retriever and generator by 8.88, 3.09, and 4.22 points on SciEGQA, ViDoRe-v3, and MMLongBench-Doc~\citep{3737916.3740957}, respectively.
\sname{} also surpasses its page-only variant under the same visual token budget and improves accuracy with two additional generators.
Because ColQwen3 and both \sname{} variants share the retriever and generator, these gains come from how pages are given to the generator.

In summary, our main contributions are:
\begin{itemize}
    \vspace{-5pt}
        \item \textbf{We identify the generator's visual input as an overlooked bottleneck in multimodal document RAG.}
        Auditing five prior methods on SciEGQA oracle pages under a common image processor, we find that they give the generator few visual tokens, whereas increasing the visual token budget and refining page content improve accuracy with the pages and generator held fixed.
        \item \textbf{We define Loud and Clear as two requirements for using evidence in retrieved pages.}
        Controlled comparisons show that visual token budget and visual fidelity independently improve accuracy, and identify full-page scaling and adding OCR of the selected regions to the page as consistent designs across benchmarks.
        \item \textbf{We propose \sname{}, a training-free framework that outperforms eight prior methods on three benchmarks.}
        Without ground-truth annotations, \sname{} outperforms its page-only variant under the same visual token budget and improves over the ColQwen3 baseline across three generators.
\end{itemize}

\clearpage
\section{Related Work}
\label{sec:related}

\subsection{Visual retrieval for multimodal document RAG}
\label{sec:related_visual_retrieval}

Recent multimodal document retrieval work has increasingly explored image-native models that preserve visual and structural information without relying on OCR-based content extraction.
DSE~\citep{ma-etal-2024-unifying} encodes each document screenshot into a single dense vector, whereas ColPali~\citep{ICLR2025_99e9e141} uses patch-level embeddings with ColBERT's~\citep{Khattab2020ColBERTEA} late-interaction scoring while keeping the page as the retrieval unit.

Retrieval and its evaluation have expanded beyond page-level relevance toward finer-grained evidence and document structure.
LILaC~\citep{yun-etal-2025-lilac} and LAD-RAG~\citep{sourati-etal-2026-lad} use document graphs for query-relevant evidence selection.
On the evaluation side, ViDoRe~\citep{ICLR2025_99e9e141} evaluates page retrieval across domains and languages, while MMDocIR~\citep{dong-etal-2025-mmdocir} adds layout-level evidence labels in long documents.
ViDoRe-v3~\citep{loison-etal-2026-vidore} broadens the scope to answer generation and visual grounding.

Multimodal document RAG has advanced largely through retrieval, improving evidence selection and evaluation.
Yet better evidence selection does not by itself establish whether the generator can use the evidence in a retrieved page, and how those pages are given to the generator remains comparatively underexplored.
This gap motivates our study of the accuracy lost through the generator's input even when the correct pages are retrieved.

\subsection[Visual token capacity in vision-language models]{Visual token capacity in vision-language models}
\label{sec:related_visual_capacity}

Modern VLMs use dynamic resolution and image tiling, so the number of visual tokens and their visual fidelity vary with the input~\citep{wang2024qwen2vlenhancingvisionlanguagemodels,liu2024llavanext,chen2024fargpt4vclosinggap}.
In multimodal document RAG, the visual fidelity of retrieved pages depends on how they are resized and encoded as visual tokens.
Because processing more visual tokens increases inference cost, many methods improve the accuracy-cost trade-off by reducing or reallocating tokens.

For token pruning, FastV~\citep{10.1007/978-3-031-73004-7_2} and SparseVLM~\citep{pmlr-v267-zhang25s} use attention-based pruning, with SparseVLM incorporating text guidance.
VisionZip~\citep{Yang_2025_CVPR} and VisionTrim~\citep{yu2026visiontrim} combine token selection and merging.
Recent analysis shows that the layer depth at which visual tokens remain informative varies with the task's visual demands and model capacity~\citep{Wang_2026_CVPR}.
For budget allocation, HiRED~\citep{10.1609/aaai.v39i2.32171} uses attention to allocate a fixed token budget across a full-image view and high-resolution tiles obtained by partitioning the same input image into a regular spatial grid.
PromPrune~\citep{lee2026balancingsaliencycoveragesemantic} adapts the balance between locally salient and globally diverse tokens to each image.
At the layer level, Optimal Vision Token Reduction~\citep{Zhao_2025_CVPR} searches for layer-wise retention schedules to reduce computation or improve accuracy under a prescribed budget.

These studies establish that visual token selection and allocation affect the accuracy and efficiency of VLM inference.
They focus on reducing computation or allocating a prescribed visual token budget.
This focus leaves open how much accuracy a low visual token budget sacrifices in multimodal document RAG.
With the retrieved pages and generator held fixed, we vary the visual token budget and visual fidelity separately and compare alternative ways of spending a matched budget, such as spatial tiling and multi-scale views (Section~\ref{sec:loud}).
Our Loud study asks how to use additional visual token capacity to preserve evidence in retrieved pages and how much accuracy each use recovers.

\subsection[Content refinement in multimodal document RAG]{Content refinement in multimodal document RAG}
\label{sec:related_evidence_presentation}

On the generator side, many multimodal document RAG systems pass retrieved page images directly to a VLM without refining their content.
M3DocRAG~\citep{Cho_2025_ICCV}, SV-RAG~\citep{ICLR2025_0d40c7b4}, and VDocRAG~\citep{Tanaka_2025_CVPR} follow this page-based design.
VisRAG~\citep{ICLR2025_3640a199} compares page concatenation, weighted selection over per-page answers, and native multi-image input.
ViDoRAG~\citep{wang-etal-2025-vidorag} selects pages from coarse thumbnails, inspects the selected pages at high resolution, and uses a reference page to verify the final answer.

Other systems refine evidence before or during answer generation.
VisDoMRAG~\citep{suri-etal-2025-visdom} curates parallel textual and visual evidence and combines their reasoning outputs through modality fusion.
RegionRAG~\citep{li2026regionrag} replaces full-page inputs to the generator with retrieved query-relevant semantic regions.
MDocRAG-RL~\citep{wang-2026-mdocrag} lets the generator crop selected regions from retrieved pages and add the re-encoded regions as new visual observations.
Related work in text RAG also processes retrieved context before generation, summarizing retrieved documents~\citep{xu2024recomp} or removing prompt tokens~\citep{jiang2024longllmlingua}.

These multimodal systems show that generation outcomes can depend on how retrieved pages are given to the generator.
However, gains within these different pipelines do not by themselves establish which content refinement helps the generator use the same retrieved pages under a fixed visual token budget.
We therefore compare content refinements of the same retrieved pages with the generator and visual token budget held fixed (Section~\ref{sec:clear}), and use the findings to develop our Clear design.

\begin{figure}[t]
    \centering
    \includegraphics[width=\textwidth]{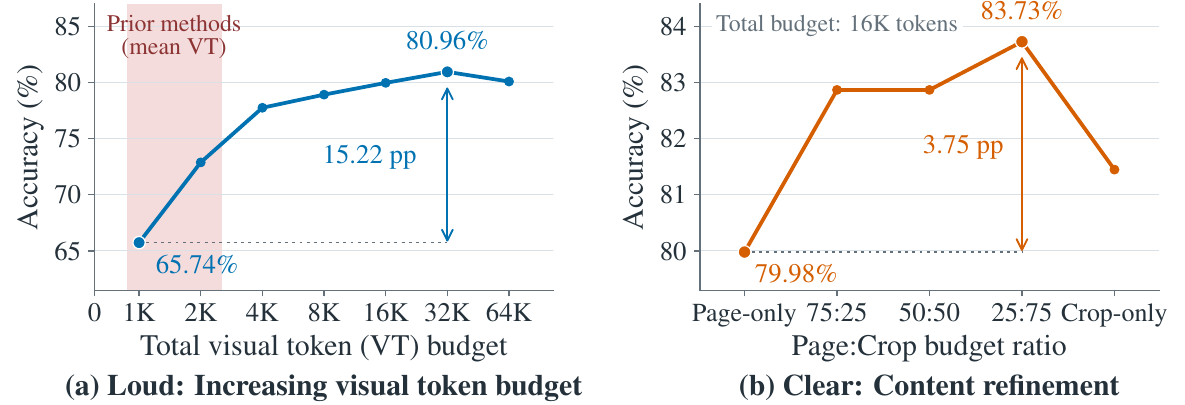}
    \vspace{-17pt}
    \caption{\textbf{Even with correct pages, larger visual token budgets and content refinement raise accuracy.}
    We diagnose this on SciEGQA with oracle pages and fixed generation and evaluation settings.
    \textbf{(a) Loud:} The red band spans five prior methods' mean usage under the Qwen3.6 image processor (0.72K--2.68K visual tokens per query).
    Increasing the page-only budget from 1K to 32K adds 15.22 pp.
    The x-axis is piecewise linear.
    \textbf{(b) Clear:} At a fixed 16K visual token budget, the best observed mixture of pages and ground-truth-matched crops adds 3.75 pp over page-only input.}
    \label{fig:post_retrieval_grid}
    \vspace{-5mm}
\end{figure}

\section[\llname{}: Improving Evidence Use in Retrieved Pages]{\llname{}: Improving Evidence Use in Retrieved Pages}
\label{sec:post_retrieval_analysis}

This section asks what the generator needs from retrieved pages, holding the pages and the generator fixed.
In Sections~\ref{sec:evidence_diagnosis} to~\ref{sec:clear}, the retrieved pages of each query are its oracle pages, the ground-truth pages that contain the evidence.
Section~\ref{sec:evidence_diagnosis} diagnoses how evidence is underused, Sections~\ref{sec:loud} and~\ref{sec:clear} compare implementations of Loud and Clear, and Section~\ref{sec:loud_clear} combines them into \sname{}.

\subsection[Evidence in retrieved pages remains underused]{Evidence in retrieved pages remains underused}
\label{sec:evidence_diagnosis}

\begin{figure}[t]
    \centering
    \includegraphics[width=\linewidth]{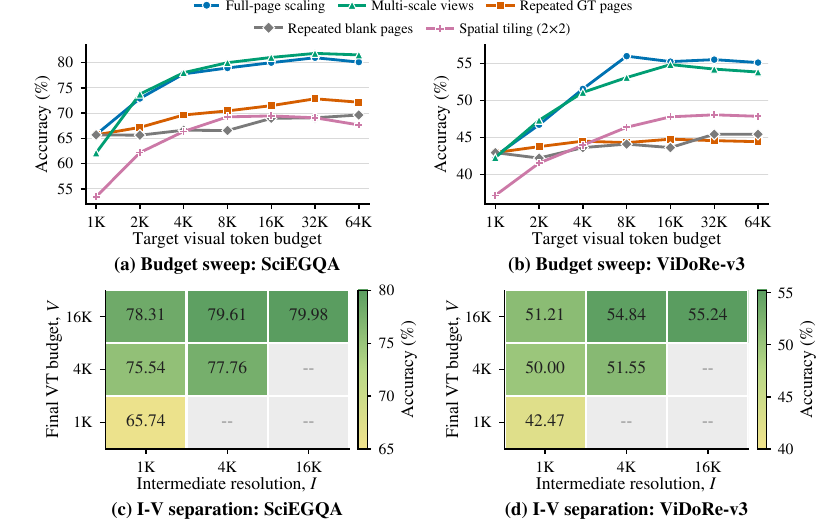}
    \vspace{-20pt}
    \caption{\textbf{How the extra tokens are spent matters, and full-page scaling spends them most consistently.}
Panels (a) and (b) show that full-page scaling outperforms repetition and spatial tiling at 2K--64K on both benchmarks, while the advantage of multi-scale views varies across benchmarks.
Panels (c) and (d) show that visual fidelity, controlled by intermediate resolution $I$, and the final visual token budget $V$ each improve accuracy with the other held fixed under two-stage resizing.
Budgets sum visual tokens across all input images, and 1K denotes 1,000 tokens.
Gray cells with $I>V$ are untested.
Color scales are independently normalized by benchmark.}
    \vspace{-14pt}
    \label{fig:section32-loud-analysis}
\end{figure}

To test how well the generator uses evidence in retrieved pages, we run a controlled diagnosis on SciEGQA~\citep{yu2026sciegqadatasetscientificevidencegrounded}, which provides ground-truth evidence boxes.
We vary the visual token budget and content refinement while fixing the retrieved pages, generator, prompt, decoding, and evaluation.
We adjust the spatial resolution of the input views to target a visual token budget across all images per query, as detailed in Appendix~\ref{app:setup}.

Prior document RAG methods leave much of the generator's visual token capacity unused.
Specifically, five prior methods give the generator 0.72K--2.68K visual tokens per query on average on the same retrieved pages when adapted to the Qwen3.6 image processor, as detailed in Appendix~\ref{app:prior_input_policies}.
However, Figure~\ref{fig:post_retrieval_grid}(a) shows that increasing the visual token budget from 1K to 32K with page-only inputs improves accuracy from 65.74\% to 80.96\%, a gain of 15.22 points.
Because the retrieved pages and generator are unchanged, this gain shows that a low visual token budget can leave available evidence underused.
These results motivate \textbf{Loud}, making full use of visual token capacity.

Moreover, content refinement can further improve generation accuracy at a fixed visual token budget.
At 16K, the best observed mixture, assigning 25\% of the budget to full-page views and 75\% to ground-truth-matched crops (ground-truth crops for short), achieves 83.73\% accuracy compared with 79.98\% for page-only input, a gain of 3.75 percentage points (Figure~\ref{fig:post_retrieval_grid}(b)).
This gain motivates \textbf{Clear}, refining the page content so that the evidence it already contains becomes prominent.

\subsection[Loud: Making full use of visual token capacity]{Loud: Making full use of visual token capacity}
\label{sec:loud}

To determine how to make full use of visual token capacity, we compare five ways of increasing the visual token budget (Appendix~\ref{app:loud}).
On SciEGQA (1,623 queries) and five public English-language datasets from ViDoRe-v3 (1,488 queries), we hold the retrieved pages and generation and evaluation settings fixed.
Each budget is a query-level target for total visual tokens produced by the generator's image processor across all input images, excluding the question and instruction text.

The first two test whether a larger budget helps without raising visual fidelity.
\textbf{Repeated blank pages} add blank images while leaving the page views unchanged. 
\textbf{Repeated pages} instead allocate additional tokens to repeated views of the same pages.
The remaining conditions test how the resolution and organization of page views affect the benefit of additional tokens.
\textbf{Full-page resolution scaling} (full-page scaling for short) increases the resolution of each page.
\textbf{Multi-scale views} add full-page views at multiple resolutions.
\textbf{Spatial tiling} uses four uniform $2\times2$ tiles per page as separate input views, without retaining a full-page view.

How the additional tokens are spent matters (Figure~\ref{fig:section32-loud-analysis}(a) and (b)).
Full-page resolution scaling outperforms repeated pages and repeated blank pages at 2K--64K, and spatial tiling at every budget on both benchmarks.
Multi-scale views outperform full-page scaling on SciEGQA at 2K--64K, whereas on ViDoRe-v3 they lead only at 2K and full-page scaling performs better at 4K--64K.

\begin{table}[t]
    \centering
    \setlength{\belowcaptionskip}{\abovecaptionskip}
    \setlength{\abovecaptionskip}{0pt}
    \caption{\textbf{At a fixed budget, suppressing surrounding content and adding selected-region OCR to page images both beat page-only input.}
    Accuracy (\%) is measured using the same retrieved pages and generator.
    Page + Crop allocates 4K VT to pages and 12K to crops.
    Blue marks selected regions and gray marks surrounding content.
    Bold marks the best per dataset and refinement group.}
    \label{tab:clear-presentation}
    \vspace{-5pt}
    \begingroup
    \small
    \setlength{\tabcolsep}{3pt}
    \renewcommand{\arraystretch}{1.18}
    \newlength{\clearlastwidth}
    \setlength{\clearlastwidth}{\dimexpr\linewidth-45pt-34pt-47pt-34pt-38pt-51pt-43pt-44pt\relax}
    \newcommand{\clearhead}[2]{\parbox[c][26pt][c]{#1}{\centering\bfseries #2}}
    \newcommand{\clearbudget}[2]{\parbox[c][12pt][c]{#1}{\centering #2}}
    \newcommand{\clearicon}[2]{\parbox[c][43pt][c]{#1}{\centering\includegraphics[width=\dimexpr#1-2pt\relax,height=39pt,keepaspectratio]{attachments/clear_input_#2.pdf}}}
    \begin{tabular}{@{}m{45pt}>{\centering\arraybackslash}m{34pt}>{\centering\arraybackslash}m{47pt}>{\centering\arraybackslash}m{34pt}>{\centering\arraybackslash}m{38pt}>{\centering\arraybackslash}m{51pt}@{\hspace{8pt}}>{\centering\arraybackslash}m{43pt}>{\centering\arraybackslash}m{\clearlastwidth}@{}}
        \toprule
        \multicolumn{8}{@{}l@{}}{\hspace*{51pt}\makebox[228pt]{\fontsize{9}{10.8}\selectfont\textit{Visual refinements}}\hspace{8pt}\makebox[\dimexpr43pt+6pt+\clearlastwidth\relax]{\fontsize{9}{10.8}\selectfont\textit{OCR-based refinements}}} \\
        \noalign{\vskip\aboverulesep\hbox{\hskip51pt\vrule width228pt height\cmidrulewidth\hskip8pt\vrule width\dimexpr43pt+6pt+\clearlastwidth\relax height\cmidrulewidth}\vskip\belowrulesep}
        \parbox[c][26pt][c]{45pt}{\textbf{Refinement}} & \clearhead{34pt}{Page} & \clearhead{47pt}{Whiteout\\Page} & \clearhead{34pt}{Crop} & \clearhead{38pt}{Canvas} & \clearhead{51pt}{Page + Crop} & \clearhead{43pt}{Crop OCR} & \clearhead{\clearlastwidth}{Page +\\\mbox{Crop OCR}} \\
        \midrule
        \parbox[c][43pt][c]{45pt}{Input view} & \clearicon{34pt}{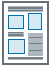} & \clearicon{47pt}{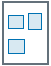} & \clearicon{34pt}{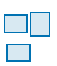} & \clearicon{38pt}{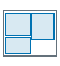} & \clearicon{51pt}{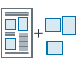} & \clearicon{43pt}{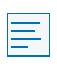} & \clearicon{\clearlastwidth}{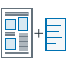} \\
        \parbox[c][12pt][c]{45pt}{Budget} & \clearbudget{34pt}{\fontsize{8}{9.6}\selectfont 16K VT} & \clearbudget{47pt}{\fontsize{8}{9.6}\selectfont 16K VT} & \clearbudget{34pt}{\fontsize{8}{9.6}\selectfont 16K VT} & \clearbudget{38pt}{\fontsize{8}{9.6}\selectfont 16K VT} & \clearbudget{51pt}{\fontsize{8}{9.6}\selectfont 16K VT} & \clearbudget{43pt}{\fontsize{8}{9.6}\selectfont\mbox{OCR $\leq 16$K}} & \clearbudget{\clearlastwidth}{\fontsize{8}{9.6}\selectfont\mbox{OCR+VT $\leq 16$K}} \\
        \midrule
        SciEGQA & 79.98 & 82.01 & 81.45 & 81.15 & \textbf{83.73} & 53.97 & \textbf{82.87} \\
        ViDoRe-v3 & 55.24 & \textbf{56.05} & 51.55 & 55.65 & 52.62 & 53.16 & \textbf{56.12} \\
        \bottomrule
    \end{tabular}
    \endgroup
    \vspace{-15pt}
\end{table}

Because full-page resolution scaling raises both the visual fidelity of the input and the final visual token budget, we separate the two with a two-stage resizing procedure (Appendix~\ref{app:loud-two-stage}).
We first resize each page to an intermediate resolution $I$, expressed in token-equivalent units, which controls visual fidelity before final resizing.
We then resize this intermediate image to a target final visual token budget $V$.
Holding $I$ fixed while increasing $V$ enlarges the same intermediate image without raising visual fidelity or recovering information lost in the first resize.

Visual fidelity and the final visual token budget each improve accuracy when the other is held fixed (Figure~\ref{fig:section32-loud-analysis}(c) and (d)).
With $V$ fixed at 4K or 16K, accuracy increases monotonically as the intermediate resolution $I$ increases on both benchmarks.
Likewise, with $I$ fixed at 1K or 4K, accuracy increases monotonically as the final visual token budget $V$ increases, although the same intermediate image is only enlarged.
Thus, higher visual fidelity alone does not explain the gains: a larger visual token budget can improve the use of evidence already preserved in the intermediate image.

We adopt full-page resolution scaling for Loud.
It increases both visual fidelity and the visual token budget while outperforming repetition and spatial tiling.
Because the advantage of multi-scale views varies across benchmarks, we use a single scaled view per page as a simple common choice.

\begin{tcolorbox}[colback=gray!5,colframe=gray!50,boxrule=0.5pt,arc=2pt,
    left=6pt,right=6pt,top=2pt,bottom=2pt,before skip=4pt,after skip=0pt]
    \textbf{Finding 1. Loud}\par\smallskip
    \noindent Allocating more visual tokens can improve the use of the same evidence even at fixed visual fidelity, and at a fixed budget, higher visual fidelity improves it further.
    The benefit of additional tokens depends on how they are spent.
\end{tcolorbox}

\subsection[Clear: Making evidence prominent]{Clear: Making evidence prominent}
\label{sec:clear}
\textbf{Loud} addresses visual token capacity, but leaves open how to make answer-relevant content prominent within the retrieved pages.
We compare content refinements that suppress surrounding content with those that present the selected regions separately, either as images or as text.
Using the SciEGQA and ViDoRe-v3 English cohorts and the fixed settings of Section~\ref{sec:loud}, we select regions by matching MinerU~\citep{wang2024mineruopensourcesolutionprecise} outputs to ground-truth evidence boxes.
Table~\ref{tab:clear-presentation} compares content refinements at a 16K visual token budget (Appendix~\ref{app:clear-construction} provides construction details, additional comparisons, and uncertainty estimates).
For conditions with OCR, the visual token budget includes OCR text tokens and visual tokens.
In tables, VT denotes visual tokens from images.

\begin{figure}[t]
    \centering
    \includegraphics{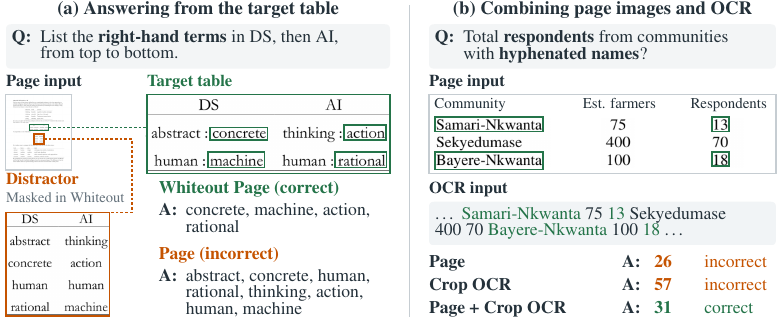}
    \vspace{-8pt}
    \caption{\textbf{Content refinement corrects answers whose evidence is already in the retrieved pages.}
    \textbf{(a)} Whiteout Page masks the distractor table and corrects the answer without enlarging selected regions or changing VT.
    \textbf{(b)} Only Page + Crop OCR gives the correct total, 31 ($13+18$).
    For readability, table excerpts are enlarged and annotated, and OCR text is rewrapped without reordering.}
    \label{fig:clear-qualitative}
    \vspace{-15pt}
\end{figure}

Suppressing surrounding content offers a direct way to make the selected regions prominent within the page.
Whiteout Page (Whiteout for short) replaces pixels outside the selected regions with white, preserving the selected pixels at their original positions, the page dimensions, and visual token budget.
Whiteout improves accuracy over Page by 2.03 percentage points on SciEGQA and 0.81 on ViDoRe-v3 without enlarging the selected regions.
The observed gains are larger when the selected regions occupy a smaller fraction of the pages (Table~\ref{tab:clear-coverage}).
Figure~\ref{fig:clear-qualitative}(a) illustrates Whiteout correcting an error where Page lists terms from the other table rather than the target table.
Whiteout removes the other table and yields the correct terms from the target table.

Extracting or regrouping the selected regions offers no consistent advantage over keeping them in their original positions.
We present the regions as separate images (Crop), group them into a canvas per page (Canvas), or pair them with full-page views (Page + Crop).
Crop underperforms Whiteout on both benchmarks, although Canvas closes much of the gap on ViDoRe-v3.
Page + Crop is the best visual refinement on SciEGQA but falls below Page alone on ViDoRe-v3.
Inspected cases show that the original page keeps heading associations and captions that refined inputs can lose (Figure~\ref{fig:clear-context}).

Adding OCR of the selected regions to page images outperforms either input alone.
Crops present the selected regions as images, leaving the generator to read the text within them.
We therefore examine whether making this text explicit through OCR helps the generator use the selected regions.
Crop OCR supplies this text alone, whereas Page + Crop OCR combines it with page images, allocating the 16K visual token budget to OCR first and the remainder to page images.
OCR alone scores below Page on both benchmarks, but their combination exceeds both, improving over Page by 2.89 percentage points on SciEGQA and 0.88 on ViDoRe-v3.
The ViDoRe-v3 gain is small, and its paired bootstrap interval includes zero.
Page + Crop OCR also scores above Page + Crop on ViDoRe-v3 and slightly below it on SciEGQA.
Figure~\ref{fig:clear-qualitative}(b) illustrates how selected text can complement page images.
OCR provides the relevant community names and counts in a continuous text sequence, while the page retains their row and column layout.
Page and Crop OCR return incorrect respondent totals of 26 and 57, respectively, whereas their combination gives the correct total of 31.

We adopt Page + Crop OCR as our Clear design, making selected text explicit while retaining the pages as a visual reference.
Both Whiteout Page and Page + Crop OCR improve over Page on the two benchmarks.
However, with identical OCR text, Whiteout and Canvas each raise accuracy on ViDoRe-v3 but lower it on SciEGQA relative to Page + Crop OCR (Table~\ref{tab:clear-matched-ocr}).
Neither offers a consistent advantage, so we retain the simpler Page + Crop OCR input.

\begin{tcolorbox}[colback=gray!5,colframe=gray!50,boxrule=0.5pt,arc=2pt,
    left=6pt,right=6pt,top=2pt,bottom=2pt,before skip=4pt,after skip=0pt]
    \textbf{Finding 2. Clear}\par\smallskip
    \noindent Content refinement that makes evidence prominent can improve the use of information already present in the retrieved pages.
    Making selected text explicit while retaining page images is one effective way to do so, with the combined input outperforming either input alone.
\end{tcolorbox}

\subsection{The \sname{} framework}
\label{sec:loud_clear}

\sname{} applies the Loud and Clear designs to end-to-end RAG by combining scaled retrieved pages with OCR from query-relevant regions.
Without additional training, \sname{} reuses the retriever's token embeddings to select regions and changes only how the retrieved pages are given to the generator.
Following the \textbf{Loud} design, we use full-page scaling to keep visual fidelity high within an expanded visual token budget.
Within this budget, we combine page images with OCR text from selected regions, applying the \textbf{Clear} design so that explicit text complements rather than replaces page images.

\begin{table}[t]
    \centering
    \setlength{\belowcaptionskip}{\abovecaptionskip}
    \setlength{\abovecaptionskip}{0pt}
    \caption{\textbf{\sname{} reaches the highest accuracy on all three benchmarks.} The first three methods use their released generator configurations. All remaining methods use Qwen3.6-35B-A3B. ColQwen3 and both \sname{} variants share the retriever and generator, and \sname{} w/o Clear gives the whole budget to page images. The margin over ColQwen3, the strongest prior baseline on each benchmark, is 8.88, 3.09, and 4.22 percentage points on SciEGQA, ViDoRe-v3, and MMLongBench-Doc.}
    \vspace{-5pt}
    \label{tab:rag-top4}
    \begingroup
    \small
    \setlength{\tabcolsep}{4pt}
    \renewcommand{\arraystretch}{1.0}
    \begin{tabular}{@{}>{\raggedright\arraybackslash}m{.40\linewidth}*{2}{>{\centering\arraybackslash}m{.155\linewidth}}>{\centering\arraybackslash}m{\dimexpr.29\linewidth-6\tabcolsep\relax}@{}}
        \toprule
        \textbf{Method} & \textbf{SciEGQA} & \textbf{ViDoRe-v3} & \textbf{MMLongBench-Doc} \\
        \midrule
        \mbox{VDocRAG \citep{Tanaka_2025_CVPR}} & 16.14 & 23.64 & 19.07 \\[1pt]
        \mbox{SV-RAG-InternVL2 \citep{ICLR2025_0d40c7b4}} & 16.94 & 24.58 & 19.98 \\[1pt]
        \mbox{EVisRAG-7B \citep{sun2026visrag20mitigatingvisualhallucinations}} & 20.46 & 23.10 & 32.63 \\
        \midrule
        \mbox{RegionRAG \citep{li2026regionrag}} & 28.77 & 32.10 & 24.38 \\[1pt]
        \mbox{ViSAR \citep{mialland2026visartrainingfreeadaptivekretrieval}} & 60.20 & 43.32 & 39.51 \\[1pt]
        \mbox{ColPali \citep{ICLR2025_99e9e141}} & 64.08 & 41.64 & 39.78 \\[1pt]
        \mbox{M3DocRAG \citep{Cho_2025_ICCV}} & 67.65 & 43.32 & 41.34 \\
        \midrule
        \mbox{ColQwen3 \footnotesize (baseline) \citep{huang2025beyond}} & 69.19 & 46.61 & 41.61 \\
        \sname{} w/o Clear & 75.48 & 49.16 & 43.54 \\
        \textbf{\sname{}} & \textbf{78.07} & \textbf{49.70} & \textbf{45.83} \\
        \bottomrule
    \end{tabular}
    \endgroup
    \vspace{-10pt}
\end{table}

To form this input, \sname{} selects query-relevant regions in the top-$k$ pages from our baseline retriever, ColQwen3~\citep{huang2025beyond}.
For each page, we map MinerU's region bounding boxes~\citep{wang2024mineruopensourcesolutionprecise} to the page embedding's visual token grid.
We then score each region using ColBERT's MaxSim-based late interaction~\citep{Khattab2020ColBERTEA}, summing each query token's maximum similarity to visual tokens within that region.
For OCR, we select each page's top-$m$ regions by score, or all detected regions if fewer than $m$.
All retrieved pages remain as visual context.

\sname{} then fits selected-region OCR and page images into one expanded visual token budget.
We extract OCR text at the original page resolution as in Section~\ref{sec:clear} and group it by page in reading order.
We set the visual token budget for actual retrieval to match the token expansion used by the Page + Crop OCR condition, which is about 5.12 times that of the same oracle pages at the baseline input resolution on average across the two benchmarks.
To approximate this token expansion in actual retrieval, we scale the baseline input width and height by $s=2.25\approx\sqrt{5.12}$ and using the resulting total visual token count as each query's visual token budget.
As in Section~\ref{sec:clear}, we allocate this budget to OCR text first and resize the original page images to fit the remainder.
The generator then receives the page images and their OCR text, followed by the QA instruction and question.

\section[Applying \sname{} to End-to-End RAG Systems]{Applying \sname{} to End-to-End RAG Systems}
\label{sec:rag_evaluation}

We use \sname{} to evaluate whether the Loud and Clear designs selected in Section~\ref{sec:post_retrieval_analysis} remain effective in actual retrieval settings.
We compare end-to-end performance with prior RAG systems, examine the effects of visual token budgets and content refinement, and test these designs across generators.

\paragraph{Setup.}
\label{sec:rag_setup}
We evaluate on SciEGQA~\citep{yu2026sciegqadatasetscientificevidencegrounded} (1,623 queries), ViDoRe-v3~\citep{loison-etal-2026-vidore} (1,489 English queries across all five public English-language datasets), and MMLongBench-Doc~\citep{3737916.3740957} (1,091 queries).
For SciEGQA and MMLongBench-Doc, retrieval covers all pages of the document provided with each question.
For ViDoRe-v3, retrieval covers the full corpus of the corresponding domain.
We use Qwen3.6-35B-A3B as the default generator with a fixed QA instruction, zero temperature, and disabled thinking.
For \sname{}, we use $k=4$ retrieved pages, $m=5$ regions per page, and the default scale $s=2.25$ described in Section~\ref{sec:loud_clear}.
We report answer accuracy under a common evaluation protocol, and Appendix~\ref{app:setup} specifies the QA instruction, the judge model and prompt, and the configuration of each compared method.

\paragraph{End-to-end RAG performance.}
\label{sec:rag_performance}
\sname{} achieves the highest accuracy on all three benchmarks (Table~\ref{tab:rag-top4}).
We use the released generator configurations for three prior methods and the common generator for the remaining five.
In its released setting, RegionRAG selects four regions across all candidate pages, giving the generator less page content than four pages.
ColQwen3 and \sname{} use the same retrieved pages, page order, and generator.
Relative to ColQwen3, the strongest prior baseline on each benchmark, \sname{} improves accuracy by 8.88, 3.09, and 4.22 points on SciEGQA, ViDoRe-v3, and MMLongBench-Doc, respectively.
Paired query-level bootstrap intervals for all three gains exclude zero (Appendix~\ref{app:end-to-end-analysis}).
These gains indicate that our controlled findings in Section~\ref{sec:post_retrieval_analysis} provide practical guidance for improving evidence use in end-to-end RAG.

We evaluate \sname{} w/o Clear, which omits selected-region OCR text while retaining full-page scaling and allocating the entire expanded visual token budget to page images.
Both \sname{} variants share ColQwen3's retrieved pages, page order, and generator.
With Loud alone, \sname{} w/o Clear already outperforms ColQwen3 on all three benchmarks.
Under the same visual token budget, \sname{} improves accuracy over this variant by 2.59, 0.54, and 2.29 percentage points on SciEGQA, ViDoRe-v3, and MMLongBench-Doc, respectively.
These gains show that Clear further improves evidence use.

\begin{figure}[t]
    \noindent
    \begin{minipage}[t]{0.48\textwidth}
        \vspace{0pt}
        \centering
        \includegraphics[width=\linewidth]{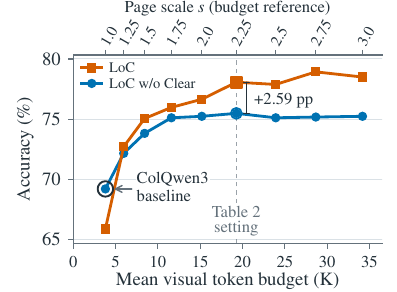}
        \vspace{-13pt}
        \setlength{\abovecaptionskip}{0pt}
        \caption{\textbf{Page scaling saturates, and Clear adds further gains.}
        SciEGQA accuracy at reference page scales $s=1.0$--$3.0$, with retrieved pages, page order, and generator fixed.
        At the smallest budget, OCR priority leaves fewer page visual tokens, possibly lowering accuracy.}
        \label{fig:rag-budget-sweep}
    \end{minipage}\hfill
    \begin{minipage}[t]{0.48\textwidth}
        \vspace{0pt}
        \centering
        \makeatletter
        \def\@captype{table}
        \makeatother
        \setlength{\belowcaptionskip}{\abovecaptionskip}
        \setlength{\abovecaptionskip}{0pt}
        \caption{\textbf{The gain over ColQwen3 carries to two additional generators.}
                ColQwen3 uses $s=1.0$ and \sname{} uses $s=2.25$. \sname{} improves over ColQwen3 on all three benchmarks with every generator we evaluate.}
        \vspace{-5pt}
        \label{tab:rag-generators}
        \begingroup
        \fontsize{9}{10.5}\selectfont
        \setlength{\tabcolsep}{2pt}
        \renewcommand{\arraystretch}{1.1}
        \newlength{\ragmetricwidth}
        \setlength{\ragmetricwidth}{\dimexpr\linewidth-64pt-6\tabcolsep\relax}
        \divide\ragmetricwidth by 3
        \begin{tabular}{@{}>{\raggedright\arraybackslash}m{64pt}*{3}{>{\centering\arraybackslash}m{\ragmetricwidth}}@{}}
            \toprule
            \textbf{Method} & {\fontsize{8}{9}\selectfont\textbf{SciEGQA}} & {\fontsize{8}{9}\selectfont\textbf{ViDoRe-v3}} & {\fontsize{8}{9}\selectfont\shortstack{\textbf{MMLong}\\\textbf{Bench-Doc}}} \\
            \midrule
            \multicolumn{4}{@{}l@{}}{{\fontsize{8}{9.5}\selectfont\textbf{Qwen2.5-VL-7B}}} \\
            ColQwen3 & 48.98 & 39.69 & 32.36 \\
            \sname{} & \textbf{56.44} & \textbf{41.91} & \textbf{36.57} \\
            \midrule
            \multicolumn{4}{@{}l@{}}{{\fontsize{8}{9.5}\selectfont\textbf{Qwen2.5-VL-32B}}} \\
            ColQwen3 & 60.01 & 47.15 & 36.57 \\
            \sname{} & \textbf{67.04} & \textbf{48.22} & \textbf{39.96} \\
            \midrule
            \multicolumn{4}{@{}l@{}}{{\fontsize{8}{9.5}\selectfont\textbf{Qwen3.6-35B-A3B}}} \\
            ColQwen3 & 69.19 & 46.61 & 41.61 \\
            \sname{} & \textbf{78.07} & \textbf{49.70} & \textbf{45.83} \\
            \bottomrule
        \end{tabular}
        \endgroup
    \end{minipage}
    \vspace{-13pt}
\end{figure}

\paragraph{Effects of visual token budget and content refinement.}
\label{sec:rag_input_design}
On SciEGQA, page scaling saturates as the budget grows, whereas Clear keeps adding gains above the smallest budget.
We sweep the reference page scale $s$ over 1.0--3.0 in 0.25 steps, fixing the retrieved pages, page order, and generator and matching the per-query visual token budget between the two variants.
Figure~\ref{fig:rag-budget-sweep} shows that \sname{} w/o Clear improves from the ColQwen3 baseline of 69.19\% at $s=1.0$ to 75.11\% at $s=1.75$, then remains near 75\% as the budget grows further.
Furthermore, \sname{} outperforms this variant at every tested scale from $s=1.25$ to $3.0$, with gains of 2.59--3.76 percentage points from $s=2.25$ (Table~\ref{tab:rag-top4}) to $s=3.0$.
Thus, combining selected-region OCR with page images adds gains even where further page scaling yields little improvement.

At the smallest budget, however, \sname{} scores 65.87\%, below the 69.19\% of \sname{} w/o Clear.
Allocating the budget to OCR first reduces the mean page visual token count from 3.80K to 1.48K, which may lower the visual fidelity of the page images that complement the selected text.
This pattern is consistent with the controlled comparisons in Section~\ref{sec:clear}, where selected-region OCR alone underperformed Page, while their combination outperformed both.
Together, these observations support keeping visual fidelity high when part of the budget goes to selected-region OCR.

\paragraph{Generalization across generators.}
\label{sec:rag_generalization}
The gains of \sname{} hold across generators.
We evaluate the Loud and Clear designs with two additional generators, Qwen2.5-VL-7B and Qwen2.5-VL-32B~\citep{bai2025qwen25vltechnicalreport}.
For each generator, ColQwen3 and \sname{} use the same retrieved pages and page order, and \sname{} retains the default scale of $s=2.25$.
Table~\ref{tab:rag-generators} shows that \sname{} improves accuracy over ColQwen3 on all three benchmarks with each generator.
These gains show that the combined \sname{} input improves evidence use with each evaluated generator.

\section[Conclusion and Discussion]{Conclusion and Discussion}
\label{sec:conclusion}

Evidence in correctly retrieved pages can remain underused by the generator.
With pages and the generator fixed, our controlled comparisons motivate Loud, sufficient visual tokens with high visual fidelity, and Clear, content refinement that makes evidence prominent.
\sname{} implements these designs through full-page scaling and selected-region OCR, without ground-truth annotations or additional training.
It outperforms eight prior methods on three benchmarks and improves over the ColQwen3 baseline across three generators.
These findings highlight the importance of how retrieved pages are presented to the generator alongside which pages are retrieved.

\paragraph{Limitations.}
LoC's larger visual token budgets introduce a trade-off between accuracy and inference latency (Appendix~\ref{app:latency}).
Clear relies on document parsing and OCR quality, and allocating tokens to OCR can reduce accuracy under tight budgets. 
Our evaluation focuses on English documents and Qwen-family generators and uses a single reference-based judge. 
Adaptive budget allocation and broader evaluation across languages, model families, and judges are directions for future work.

\subsection*{Acknowledgements}
This work was partly supported by the Institute of Information \& Communications Technology Planning \& Evaluation (IITP) grant funded by the Korean government (MSIT) (No. RS-2026-25524173, Ultra-Long-Term Hierarchical Memory and Reasoning Architecture for Next-Generation Omnimodal Agents; 50\%) and the IITP grant for the AI Computing Support Project for R\&D funded by the Korean government (MSIT) (No. RS-2026-25505492, High-Performance Research AI Computing Infrastructure Support at the 2 PFLOPS Scale; 50\%).

\subsection*{AI use statement}
We used generative AI tools to assist with manuscript drafting and editing, literature search, code development, figure and table preparation, and interpretation of experimental results. The authors reviewed and verified all AI-assisted material and take full responsibility for the final content of the paper.

\subsection*{Reproducibility statement}
We provide the controlled experimental settings and input construction details in Appendices~\ref{app:loud}, \ref{app:clear-construction}, and~\ref{app:setup}.
The answer evaluation protocol, including the full judge prompt, and the end-to-end RAG configurations are described in Appendix~\ref{app:setup}.

\bibliography{references}
\bibliographystyle{iclr2027_conference}

\clearpage
\appendix
\etocdepthtag.toc{mtappendix}
\hypersetup{linkcolor=black}
\etocsettagdepth{mtchapter}{none}
\etocsettagdepth{mtappendix}{subsection}
\tableofcontents
\hypersetup{linkcolor=linkcol}
\clearpage
\section[Underused Visual Token Capacity in Prior Methods]{Underused Visual Token Capacity in Prior Methods}
\label{app:prior_input_policies}

We measure the visual token usage of prior document RAG methods on the SciEGQA pages used in Section~\ref{sec:post_retrieval_analysis}.
We adapt each method's image resizing and slicing settings to the common image processor used in our controlled experiments, allowing their input sizes to be measured on the same pages and token grid.
Table~\ref{tab:prior_input_vt} reports the resulting usage.
The means across the five generator-side methods range from 718.08 to 2,679.87 visual tokens per query, giving the range of 0.72K--2.68K reported in Section~\ref{sec:evidence_diagnosis}, which the abstract and introduction round to 0.7K--2.7K.
The table also reports the median and range across individual queries.

\begin{table}[h]
    \centering
    \setlength{\belowcaptionskip}{\abovecaptionskip}
    \setlength{\abovecaptionskip}{0pt}
        \caption{\textbf{Under a common image processor, five prior methods give the generator 0.72K--2.68K visual tokens per query on average.} Visual token usage on the 1,623 SciEGQA queries. Counts sum all views from all retrieved pages for each query. The SV-RAG row uses the later Self-Visual-RAG implementation described below. ColPali is a retriever-side setting and is excluded from the range reported in the main text.}
    \label{tab:prior_input_vt}
    \begin{tabular}{@{}lrrr@{}}
        \toprule
        Method & Mean & Median & Min--max \\
        \midrule
        M3DocRAG & 2,281.80 & 1,900 & 1,350--3,922 \\
        VDocRAG & 1,906.41 & 1,700 & 1,300--3,400 \\
        VisRAG & 718.08 & 630 & 486--1,260 \\
        SV-RAG (later implementation) & 2,679.87 & 2,548 & 1,372--5,096 \\
        VisDoM & 740.65 & 616 & 600--1,260 \\
        \midrule
        ColPali (retrieval) & 234.40 & 196 & 196--392 \\
        \bottomrule
    \end{tabular}
\end{table}

\paragraph{Common measurement setting.}
The audit uses the same 1,623 queries and their retrieved pages as Appendix~\ref{app:controlled_construction}.
Each query has one or two retrieved pages, with 1,941 page occurrences and 837 unique pages in total.
We start from the original page images rendered at 300 DPI, before the resizing used in the controlled budget sweep.
We fix the queries, page identities and order, original images, and final image processor while varying the method's settings.
Each method's settings are applied to all retrieved pages for a query, without applying its original retrieval top-$k$ or matching a query-level budget from the controlled sweep.
We run the processor on CPU without loading model weights, performing retrieval, or generating answers.

The processor uses the same Qwen3.6-35B-A3B checkpoint revision as Appendix~\ref{app:controlled_construction}, with Transformers~\citep{wolf-etal-2020-transformers} 5.12.1 and PyTorch~\citep{NEURIPS2019_bdbca288} 2.11.0.
Its \texttt{Qwen2VLImageProcessor}, accessed through \texttt{Qwen3VLProcessor}, uses $16\times16$ patches and a $2\times2$ spatial merge.
A processed image view of height $H$ and width $W$ therefore contributes $(H/32)(W/32)$ visual tokens.
We count the expanded image tokens and check them against the image grids, summing across all views and pages for each query before averaging over queries.
We exclude question text, image delimiters, and sequence padding but include padding encoded within an image view.
Processor measurements are reused for views with the same dimensions and area constraints, since these determine the token count.

\paragraph{Method selection and adaptation.}
We use the public settings associated with each system's default or main model setting.
The model names below identify the source of each setting.
The common processor has default minimum and maximum areas of 65,536 and 16,777,216 pixels per view, with the method-specific overrides given below.

\begin{itemize}
    \item \textbf{M3DocRAG.} We use the Qwen2-VL-7B settings~\citep{Cho_2025_ICCV}, constructing one view per page with dimensions corresponding to 144 DPI and the most frequent page dimensions in the full document.
    The document-level dimensions are determined from all pages, including those outside the retrieved set.
    We resize the original raster images to these dimensions and retain the original area limits of 3,136 and 12,845,056 pixels per view.
    This matches the target dimensions and token counts of the rendering setting without assuming pixel-identical PDF rerendering.

    \item \textbf{VDocRAG.} We use the Phi-3-V settings~\citep{Tanaka_2025_CVPR} with \texttt{num\_crops=16}, the checkpoint setting used without a generator override.
    Each page produces one global view and up to 16 local tiles.
    The original $336\times336$ views are mapped to $320\times320$ views on the common processor grid, yielding 100 tokens per view.
    Only the actual views are counted, excluding empty batch crop slots and any duplicate of the tiled canvas.

    \item \textbf{VisRAG.} We use the MiniCPM-V 2.6 multi-image settings~\citep{ICLR2025_3640a199}, retaining its slicing procedure with a base scale of 448 and patch size of 14.
    Each page produces one global view and up to nine local views.
    We represent the native resampler's fixed 64 tokens per view by a 64-token cap under the common processor, with area limits of 1,024 and 65,536 pixels per view.
    The resulting common count can be below 64 for a view, and is at most 640 per page.
    This adaptation imposes a token-count constraint and does not reproduce the learned resampler or its information preservation.

    \item \textbf{SV-RAG.} We use the InternVL2-4B settings from the later, author-linked Self-Visual-RAG implementation of SV-RAG~\citep{ICLR2025_0d40c7b4}.\footnote{The author's \href{https://jian-chen.name/posts/SV-RAG/}{project page} links to the \href{https://github.com/puar-playground/Self-Visual-RAG}{Self-Visual-RAG implementation}.}
    Its answer-generation call sets \texttt{max\_num=20}, yielding up to 20 local $448\times448$ tiles and an additional thumbnail when more than one local tile is present.
    Each tile or thumbnail contributes 196 tokens under the common processor.
    The tile limit belongs to this later implementation and is not a verified setting of the original paper.

    \item \textbf{VisDoM.} We use the Qwen2-VL-7B visual-branch settings~\citep{suri-etal-2025-visdom}, constructing one full-page view with minimum and maximum areas of 262,144 and 655,360 pixels.
    These correspond to 256--640 visual tokens per page on the common grid.
    We count the visual branch once per page, without adding another image pass for the text branch or modality fusion.
\end{itemize}

\paragraph{Retriever-side reference.}
For ColPali~\citep{ICLR2025_99e9e141}, we resize each page to $448\times448$ without preserving its aspect ratio, yielding 196 tokens per page under the common processor.
Its query-level count in Table~\ref{tab:prior_input_vt} sums these page counts over the same retrieved pages.
This row describes a retriever-side configuration, so it is kept separate from the five generator-side methods used to obtain the 0.72K--2.68K range in the main text.

\paragraph{Interpretation.}
These measurements characterize the adapted settings under a shared processor, rather than the native token counts or QA performance of the original systems.
Mapping each method's settings to a common token grid does not preserve their original resolution, token count, and information content simultaneously, particularly when a learned compression module is replaced by an input-area constraint.
The usage measurements establish the input levels of the selected methods on these pages.
The separate controlled results in Figure~\ref{fig:post_retrieval_grid}(a) establish the utility of greater visual token capacity in our setting.
The audit alone does not attribute a performance loss to any prior system as it was originally designed and trained.

\clearpage
\section[Extended Analysis of Loud]{Extended Analysis of Loud}
\label{app:loud}

This appendix details the conditions, visual token accounting, and two-stage resizing procedure used in Section~\ref{sec:loud}.
The datasets, generation settings, and answer evaluation protocol shared by all conditions are specified in Appendices~\ref{app:controlled_construction} and~\ref{app:answer_evaluation}.

\subsection{Conditions}
\label{app:loud-constructions}

We keep the oracle pages fixed and vary their visual representation to the generator.
Each query is assigned a target budget
$B\in\{1,2,4,8,16,32,64\}\mathrm{K}$, where $1\mathrm{K}=1{,}000$ tokens.
The target applies to the total visual token count across all input views of the query.
These conditions use only page images, without crops, OCR text, or other additional inputs.

\paragraph{Full-page resolution scaling.}
We represent each page with a single full-page view and adjust its resolution according to the query-level target budget.
At each budget, we resize directly from the original page image, rather than from a lower-budget input.

\paragraph{Repeated pages and repeated blank pages.}
Repeated pages uses multiple views of each retrieved page, whereas repeated blank pages combines page views with additional blank images.
The target visual token budget applies to all image views of a query.
These conditions compare repetition and blank-image addition with full-page resolution scaling at the same target budgets.

\paragraph{Multi-scale views.}
For a query with $P$ oracle pages, we construct exactly three full-page views for each page, yielding $3P$ input images at every target budget.
The three views use relative linear scales $s_\ell \in \{1,2,4\}$ and are constructed independently from the original page image.
Because visual token count scales approximately with image area, their target token shares are proportional to $s_\ell^2$, giving the ratio
$1:4:16$ across the three scales.

For a query-level target budget $B$, each page is first assigned $B/P$ tokens, and scale $\ell$ receives
\begin{equation}
b_{p,\ell}
=
\frac{B}{P}
\frac{s_\ell^2}{1^2+2^2+4^2}.
\label{eq:loud-multiscale-allocation}
\end{equation}
Thus, the three scales receive $1/21$, $4/21$, and $16/21$ of each page's target allocation.
Each view is independently quantized to the processor's 32-pixel grid, and no residual query-level correction is applied after this quantization.
Consequently, realized visual token counts can deviate from $B$, particularly for queries with many oracle pages at small budgets.
We retain all oracle pages and all three scales, ordering the views page-wise from coarse to fine, so the coarse view of each page precedes its finer views.

\paragraph{Spatial tiling.}
We divide each page into a uniform $2\times2$ grid and supply the four tiles as separate input views, without retaining a full-page view.
The tiles from all pages share the query-level target budget $B$, so each tile's resolution decreases as the number of pages grows.

\subsection{Visual token accounting and resolution allocation}
\label{app:loud-token-accounting}

The generator's image processor uses $16\times16$ pixel patches and merges $2\times2$ spatial patches into one visual token. 
For a static image whose processed dimensions are $H_j\times W_j$, both divisible by 32, the token count is
\begin{equation}
    T_j=\frac{H_jW_j}{32^2},
    \qquad
    \widehat{B}(q)=\sum_{j=1}^{M(q)}T_j,
    \label{eq:loud-token-accounting}
\end{equation}
where $M(q)$ is the number of input views for query $q$.
Equivalently, the processor grid $(t_j,h_j,w_j)$ contributes $t_jh_jw_j/4$ tokens, with $t_j=1$ for these static images.
The count includes blank image regions but excludes the question, instruction text, and image boundary markers.

For full-page resolution scaling, the reference dimensions are obtained by applying the processor's resizing rule with a per-page area ceiling of 1,003,520 pixels. 
These dimensions serve only as allocation references: the final pixels are resized directly from the original page image.
We apply a common continuous scale to the reference dimensions within a query, then map the dimensions to the processor's 32-pixel grid.
To approach the target, the allocator chooses between the resulting sizes and their adjacent grid transitions in the required direction. 
This preserves the common-scale construction up to quantization rather than assigning arbitrary independent resolutions to different pages. 
The selected combination minimizes the absolute token error among these candidates. 
Thus, $B$ is a target, and the realized count $\widehat{B}(q)$ can lie slightly above or below it.

Table~\ref{tab:loud-actual-vt} reports the mean realized visual token counts
for all five Loud constructions.
Despite processor-grid quantization, the condition-level means closely match
their target budgets across both benchmarks.

\begin{table}[h]
    \centering
    \setlength{\belowcaptionskip}{\abovecaptionskip}
    \setlength{\abovecaptionskip}{0pt}
    \caption{\textbf{All Loud constructions closely match their target visual-token budgets on average.}
    Values report the mean realized visual tokens per query.
    Across all constructions and target budgets, the largest absolute deviation of a condition mean from its target is 35.12 tokens, and the largest relative deviation is 1.32\%.}
    \label{tab:loud-actual-vt}
    \small
    \setlength{\tabcolsep}{2.5pt}
    \begin{tabular}{@{}lrrrrrrr@{}}
        \toprule
        Construction & 1K & 2K & 4K & 8K & 16K & 32K & 64K \\
        \midrule
        \multicolumn{8}{@{}l}{\textit{SciEGQA}} \\
        Full-page resolution scaling
        & 1002.8 & 1999.5 & 4009.1 & 7978.5 & 15993.2 & 32002.7 & 63977.1 \\
        Repeated pages
        & 1002.8 & 2011.9 & 3996.2 & 7998.8 & 15990.0 & 32001.6 & 64002.7 \\
        Repeated blank pages
        & 1002.8 & 2011.9 & 3996.2 & 7998.8 & 15990.0 & 32001.6 & 64002.7 \\
        Multi-scale views
        & 987.4 & 1973.6 & 4000.1 & 8016.7 & 16032.2 & 32030.5 & 64032.9 \\
        Spatial tiling ($2\times2$)
        & 997.8 & 1999.0 & 3996.2 & 8004.4 & 15994.0 & 32017.6 & 64024.3 \\
        \midrule
        \multicolumn{8}{@{}l}{\textit{ViDoRe-v3}} \\
        Full-page resolution scaling
        & 1000.0 & 2000.4 & 4000.1 & 7998.9 & 15996.9 & 32003.0 & 63997.1 \\
        Repeated pages
        & 1000.0 & 2002.5 & 3999.6 & 7998.7 & 15997.7 & 31999.7 & 63997.2 \\
        Repeated blank pages
        & 1000.0 & 2002.5 & 3999.6 & 7998.7 & 15997.7 & 31999.7 & 63997.2 \\
        Multi-scale views
        & 987.3 & 1985.0 & 3984.1 & 8006.2 & 16007.7 & 32022.9 & 64035.1 \\
        Spatial tiling ($2\times2$)
        & 999.7 & 1999.7 & 3999.6 & 7999.7 & 15998.4 & 32001.1 & 64001.5 \\
        \bottomrule
    \end{tabular}
\end{table}

At the query level, the realized budgets are also closely matched for four of the five constructions.
For full-page resolution scaling, repeated pages, repeated blank pages, and spatial tiling, no query deviates from its target by more than 5\% in any evaluated benchmark or budget.
Multi-scale views show larger deviations on ViDoRe-v3 at the smallest budgets because each page's allocation is divided across three scales and each view is independently quantized to the processor grid.
At 1K, 8.13\% (121/1,488) of ViDoRe-v3 queries deviate from the target by more than 5\%; this fraction falls to 3.02\% at 2K and 0.54\% at 4K, and is zero from 8K onward.

\subsection{Separating intermediate resolution from final budget}
\label{app:loud-two-stage}

Full-page resolution scaling changes both visual fidelity and the final visual token count. 
We separate these factors using an intermediate resolution \(I\), which controls the visual fidelity retained before final resizing, and a final budget \(V\). 
\(I\) specifies image dimensions and does not directly measure semantic information.
Let $P_i$ be an original page image, and let $d_i(b)$ denote its processed dimensions in the page-only allocation for target $b$. 
With $R(X;d)$ denoting resizing image $X$ to dimensions $d$, we construct
\begin{equation}
    Z_i^{(I)}=R(P_i;d_i(I)),
    \qquad
    X_i^{(I,V)}=R(Z_i^{(I)};d_i(V)).
    \label{eq:loud-two-stage}
\end{equation}
The 1K and 4K intermediate images are constructed independently from the original pages. 
Conditions with the same $I$ reuse the same intermediate image tensors. 
Both stages use the existing processor's bicubic resizing with antialiasing enabled, and the second resize is skipped when its input already has exactly the required final dimensions.

The final dimensions $d_i(V)$ are taken from the existing page-only allocation; they are not recomputed from the intermediate image.
Consequently, conditions with the same $V$ have identical page order, per-page final dimensions, and actual visual token counts for each query.
The input contains all page views followed by the QA instruction and question.
Reference answers are used for evaluation only and never appear in the generator input.

\paragraph{Controlled comparisons.}
At fixed $V$, increasing $I$ changes the visual fidelity retained before final resizing while preserving the final token grid. 
At fixed $I$, increasing $V$ enlarges the same intermediate image without recovering information lost in the first resize, testing whether a larger visual token budget improves the use of the preserved evidence.

We evaluate the off-diagonal conditions $(I,V)=(1\mathrm{K},4\mathrm{K})$,
$(1\mathrm{K},16\mathrm{K})$, and $(4\mathrm{K},16\mathrm{K})$.
The diagonal entries in Figure~\ref{fig:section32-loud-analysis} use the corresponding page-only results. 
Configurations with $I>V$ are outside this degrade-then-upsample design and are not evaluated.

\begin{table}[h]
    \centering
    \setlength{\belowcaptionskip}{\abovecaptionskip}
    \setlength{\abovecaptionskip}{0pt}
        \caption{\textbf{Visual fidelity and the final visual token budget each improve accuracy with the other held fixed.} Accuracy (\%) under two-stage resizing.
    Intermediate resolution controls retained source visual fidelity.
    Intermediate resolution and final visual token budget are expressed in query-level token-equivalent units, with $1\mathrm{K}=1{,}000$ tokens.
    Diagonal entries ($I=V$) reuse the page-only budget sweep, while off-diagonal entries ($I<V$) use two-stage resizing.}
    \label{tab:loud-two-stage-results}
    \small
    \begin{tabular}{@{}ccrr@{}}
        \toprule
        \shortstack{Intermediate\\resolution, $I$}
        & \shortstack{Final visual\\token budget, $V$}
        & SciEGQA & ViDoRe-v3 \\
        \midrule
            1K  & 1K  & 65.74 & 42.47 \\
            1K  & 4K  & 75.54 & 50.00 \\
            4K  & 4K  & 77.76 & 51.55 \\
            1K  & 16K & 78.31 & 51.21 \\
            4K  & 16K & 79.61 & 54.84 \\
            16K & 16K & 79.98 & 55.24 \\
        \bottomrule
    \end{tabular}
\end{table}

\subsection{Complete budget sweep}
\label{app:loud-full-results}

Table~\ref{tab:loud-full-sweep} provides the numerical values underlying the five curves in Figure~\ref{fig:section32-loud-analysis}(a) and (b). 
For full-page resolution scaling, we use the complete 1,623-query SciEGQA cohort and 1,488-query ViDoRe-v3 cohort at every budget. 
Answers reaching the 1,024-token output limit remain in the evaluation.

\begin{table}[h]
    \centering
    \setlength{\belowcaptionskip}{\abovecaptionskip}
    \setlength{\abovecaptionskip}{0pt}
    \caption{\textbf{Multi-scale views lead on SciEGQA at 2K--64K, whereas full-page resolution scaling leads on ViDoRe-v3 at 4K--64K.} 
    Accuracy (\%) across target visual token budgets for the five Loud conditions.
    Each target is summed across all input images.
    Bold indicates the highest observed accuracy at each budget within a benchmark. Values correspond to Figure~\ref{fig:section32-loud-analysis}.}
    \label{tab:loud-full-sweep}
    \begingroup
    \small
    \setlength{\tabcolsep}{3pt}
    \begin{tabular}{@{}lrrrrrrr@{}}
        \toprule
        Condition & 1K & 2K & 4K & 8K & 16K & 32K & 64K \\
        \midrule
        \multicolumn{8}{@{}l}{\textit{SciEGQA}} \\
        Full-page resolution scaling
        & \textbf{65.74} & 72.89 & 77.76 & 78.93 & 79.98 & 80.96 & 80.10 \\
        Repeated pages
        & 65.68 & 67.16 & 69.62 & 70.43 & 71.47 & 72.83 & 72.15 \\
        Repeated blank pages
        & 65.68 & 65.62 & 66.61 & 66.54 & 68.95 & 69.07 & 69.62 \\
        Multi-scale views
        & 62.11 & \textbf{73.75} & \textbf{78.00} & \textbf{79.98}
        & \textbf{81.02} & \textbf{81.82} & \textbf{81.52} \\
        Spatial tiling ($2\times2$)
        & 53.42 & 62.17 & 66.36 & 69.25 & 69.44 & 69.07 & 67.65 \\
        \midrule
        \multicolumn{8}{@{}l}{\textit{ViDoRe-v3}} \\
        Full-page resolution scaling
        & 42.47 & 46.71 & \textbf{51.55} & \textbf{55.98}
        & \textbf{55.24} & \textbf{55.51} & \textbf{55.11} \\
        Repeated pages
        & \textbf{42.94} & 43.75 & 44.49 & 44.29 & 44.76 & 44.56 & 44.42 \\
        Repeated blank pages
        & \textbf{42.94} & 42.20 & 43.62 & 44.09 & 43.62 & 45.43 & 45.43 \\
        Multi-scale views
        & 42.27 & \textbf{47.31} & 51.08 & 53.09 & 54.84 & 54.23 & 53.83 \\
        Spatial tiling ($2\times2$)
        & 37.16 & 41.53 & 43.95 & 46.37 & 47.78 & 48.05 & 47.85 \\
        \bottomrule
    \end{tabular}
    \endgroup
\end{table}

\paragraph{Results across budgets.}
Within the evaluated sweep, multi-scale views achieve higher accuracy than full-page resolution scaling on SciEGQA at 2K--64K, whereas full-page resolution scaling achieves higher accuracy on ViDoRe-v3 at 4K--64K.
Full-page resolution scaling reaches its observed maximum at 32K on SciEGQA and 8K on ViDoRe-v3.
Because the advantage of multi-scale views on SciEGQA does not carry over to ViDoRe-v3, we use a single scaled full-page view per page as a simple common choice across the two benchmarks.

\clearpage
\section[Extended Analysis of Clear]{Extended Analysis of Clear}
\label{app:clear-construction}

\subsection{Construction}
\label{app:clear-build}

\paragraph{Selected regions.}
For each MinerU region, we identify the ground-truth evidence box that maximizes the pair of intersection-over-union (IoU) and min-area coverage lexicographically.
We retain the region if either its IoU with this matched box is at least 0.5 or their intersection covers at least half of the smaller box.
We add 10 pixels of padding in a per-page reference coordinate system obtained by the Qwen image processor's smart resizing, with a maximum area of 1,003,520 pixels, and map the padded box back to the original page image.
Region matching is performed before padding; padding is applied only after a MinerU region has been selected.
The fallback to raw ground-truth boxes applies only when no matched region exists anywhere in the query.

\paragraph{Visual refinements.}
Whiteout replaces pixels outside the selected regions with white, preserving Page's image dimensions, selected pixels, page order, and visual token counts.
Crop presents the selected regions as individual images.
Page + Crop allocates 4K VT to full-page views and 12K VT to crop images.
Canvas packs the same crops separately for each page using the order-preserving rectangle-packing algorithm in the Eclipse Layout Kernel (ELK)~\citep{domros2021rectangle}, implemented in elkjs 0.12.0.
Crops are ordered by their vertical and horizontal coordinates and packed without rotation or added spacing.
We render the raw crops directly at the allocated final resolution to avoid enlarging an intermediate low-resolution canvas.
This grouping changes the original page geometry.

\paragraph{Main shared-budget OCR comparison.}
We extract text before generation resizing with the PaddleOCR components~\citep{cui2025paddleocr30technicalreport} bundled in MinerU 2.5.4, using PP-OCRv5 detection and PP-OCRv4 text recognition. 
Detected strings are concatenated in detection order and grouped by page and crop, without reconstructing table structure or correcting text using reference answers.
The main Crop OCR and Page + Crop OCR conditions in Table~\ref{tab:clear-presentation} share text extracted from the same padded crops. 
OCR-only inputs are truncated to at most 16,000 text tokens. 
Combined inputs allocate tokens to OCR first and resize the visual views into the remaining budget, enforcing $T_{\mathrm{OCR}} + V \leq 16{,}000$.
This cap includes OCR text and visual tokens, excluding the question and other prompt text. 
If OCR fills the budget, no images are included, as occurs for three ViDoRe-v3 queries in the main Page + Crop OCR condition of Table~\ref{tab:clear-presentation}.

\subsection{Additional comparisons}
\label{app:clear-comparisons}

\paragraph{Visual refinements with identical OCR.}
Table~\ref{tab:clear-matched-ocr} isolates the effect of visual refinement using identical OCR text extracted from the padded crops described in Appendix~\ref{app:clear-build}.
The Page + Crop OCR row is the same condition reported in Table~\ref{tab:clear-presentation}; the other rows change only the visual refinement.

\begin{table}[h]
    \centering
    \setlength{\belowcaptionskip}{\abovecaptionskip}
    \setlength{\abovecaptionskip}{0pt} 
    \caption{\textbf{With the same padded-crop OCR, no visual refinement leads on both benchmarks.} Accuracy with identical crop OCR and different visual inputs.
    All refinements share the same OCR text under an OCR + VT $\leq$ 16K budget.
    Bold marks the highest accuracy per dataset.}
    \label{tab:clear-matched-ocr}
    \begingroup
    \small
    \setlength{\tabcolsep}{10pt}
    \renewcommand{\arraystretch}{1.2}
    \begin{tabular}{@{}lcc@{}}
        \toprule
        & \multicolumn{2}{c@{}}{\textbf{Accuracy (\%)}} \\
        \cmidrule(l){2-3}
        \textbf{Refinement} & SciEGQA & ViDoRe-v3 \\
        \midrule
        Page + Crop OCR & \textbf{82.87} & 56.12 \\
        Whiteout Page + Crop OCR & 82.50 & 57.12 \\
        Canvas + Crop OCR & 81.45 & \textbf{57.26} \\
        \bottomrule
    \end{tabular}
    \endgroup
\end{table}

\paragraph{Adding OCR to fixed page images.}
Table~\ref{tab:clear-ocr-comparisons} tests a different question: whether selected-region OCR provides additional value when the page images themselves are left unchanged. 
We therefore preserve the 16K Page images and their visual tokens exactly, and add the padded-crop OCR as additional text. 
Unlike the shared-budget comparisons above, the OCR tokens in this setting are not counted against the
16K visual token budget of the page images.

\begin{table}[h]
    \centering
    \setlength{\belowcaptionskip}{\abovecaptionskip}
    \setlength{\abovecaptionskip}{0pt}
    \caption{\textbf{Crop OCR helps even when the page images are preserved exactly.}
    Accuracy (\%) when adding crop OCR while preserving the Page images and their VT exactly.
    OCR uses additional tokens beyond the 16K visual token budget. Accuracy increases over Page by 2.09 and 0.61 percentage points on SciEGQA and ViDoRe-v3, respectively.}
    \label{tab:clear-ocr-comparisons}
    \small
    \begin{tabular}{@{}llcc@{}}
        \toprule
        Condition & Budget & SciEGQA & ViDoRe-v3 \\
        \midrule
        Page & 16K VT & 79.98 & 55.24 \\
        Page + Crop OCR & 16K VT + OCR & 82.07 & 55.85 \\
        \bottomrule
    \end{tabular}
\end{table}

\paragraph{Paired uncertainty estimates.}
For the shared-budget comparisons in Table~\ref{tab:clear-presentation}, we report paired-query 95\% bootstrap percentile intervals using 20,000 resamples.
Each comparison uses the same 1,623 SciEGQA or 1,488 ViDoRe-v3 queries and their saved strict-correct judgments; ViDoRe-v3 accuracy is pooled over queries.
For Page$\rightarrow$Whiteout, the accuracy changes are $+2.03$ percentage points (pp), $[0.49,3.51]$, on SciEGQA and $+0.81$ pp, $[-1.14,2.69]$, on ViDoRe-v3.
For Page$\rightarrow$Page+Crop OCR, they are $+2.89$ pp, $[1.23,4.56]$, and $+0.88$ pp, $[-1.14,2.82]$, respectively.
For Crop$\rightarrow$Canvas, they are $-0.30$ pp, $[-1.54,0.86]$, and $+4.10$ pp, $[2.35,5.85]$, respectively.
Displayed changes subtract rounded accuracies, while intervals use unrounded paired scores.
The original analysis uses a shared random-number stream initialized with seed 20260915 and linear percentile interpolation.
These comparisons form an exploratory, post hoc subset of 38 dataset-specific contrasts.
The intervals are pointwise, without multiple-comparison correction, and condition on saved outputs and judgments; they do not measure generation-rerun or judge variability.

\subsection[Region selection and answer content]{Region selection and answer content}
\label{app:clear-observations}

\paragraph{Selected area and suppression.}
Table~\ref{tab:clear-coverage} groups the Page--Whiteout comparison by the fraction of page area occupied by selected regions.
The observed gains are larger in the groups with smaller selected fractions on both benchmarks.
The groups are not matched for question type, document structure, or difficulty, so this association neither isolates an effect of area nor supports a rule that chooses the input by the area of its selected regions.

\begin{table}[h]
    \centering
    \setlength{\belowcaptionskip}{\abovecaptionskip}
    \setlength{\abovecaptionskip}{0pt}
        \caption{\textbf{Whiteout gains are larger when the selected regions cover less of the page.} Page-to-Whiteout accuracy changes by selected area.
    Selected area is the sum of selected-box union areas across a query's pages divided by their total area.
    Each difference uses paired outcomes within the indicated group.
    The middle group includes 10\% and excludes 30\%. The groups are not matched in other respects, so the association is not a causal effect of area.}
    \label{tab:clear-coverage}
    \small
    \setlength{\tabcolsep}{9pt}
    \renewcommand{\arraystretch}{1.15}
    \begin{tabular}{@{}lcccc@{}}
        \toprule
        & \multicolumn{2}{c}{\textbf{SciEGQA}} & \multicolumn{2}{c@{}}{\textbf{ViDoRe-v3}} \\
        \cmidrule(lr){2-3}\cmidrule(l){4-5}
        \textbf{Selected area} & $N$ & $\Delta$ accuracy (pp) & $N$ & $\Delta$ accuracy (pp) \\
        \midrule
        $<10\%$ & 312 & $+4.49$ & 223 & $+1.79$ \\
        $10$--$30\%$ & 808 & $+2.60$ & 717 & $+1.12$ \\
        $\geq30\%$ & 503 & $-0.40$ & 548 & $0.00$ \\
        \bottomrule
    \end{tabular}
\end{table}

\paragraph{Competing passages.}
In SciEGQA query \texttt{2412.20176-p1-249}, the question asks for the numbers of economic concepts and data sources listed in the selected passage.
The page's abstract names three concepts, whereas the selected introduction passage lists four, including the Poverty Gap Index.
Both passages name three data sources.
Page answers ``3 economic concepts and 3 data sources,'' matching the abstract, while Whiteout answers ``4 economic concepts and 3 data sources,'' matching the selected passage and reference answer.
The selected passage retains identical pixels and position, and both inputs use 16,006 VT.
The answer correspondence is observable, but it does not establish where the model attended internally.

\paragraph{Selecting a table entry with OCR.}
SciEGQA query \texttt{2501.00176-p15-468} asks for the latitude of the site with the highest MLAT among rows citing \textit{The Times of India}.
Bhownuggur has MLAT 12.9 and latitude N$21^{\circ}46^{\prime}$, whereas Jacobabad has the highest qualifying MLAT, 19.9, and latitude N$28^{\circ}17^{\prime}$.
Page answers with Bhownuggur's latitude.
Crop OCR and Page + Crop OCR both return Jacobabad's latitude, matching the reference.
The OCR includes both rows' sites, coordinates, MLAT values, and references.
The combined input uses 15,012 VT, compared with 15,984 for Page.
This case illustrates a more accurate answer selection when the relevant text is supplied explicitly, without establishing that Page failed to read the individual values.
Together with the corrected respondent total in Figure~\ref{fig:clear-qualitative}(b), it illustrates how supplying selected text alongside page images can improve answers from tables.
These inspected cases illustrate possible improvements, not how often such improvements occur across all queries.

\subsection{Examples of page context}
\label{app:clear-cases}

\begin{figure}[h]
    \centering
    \includegraphics[width=\linewidth]{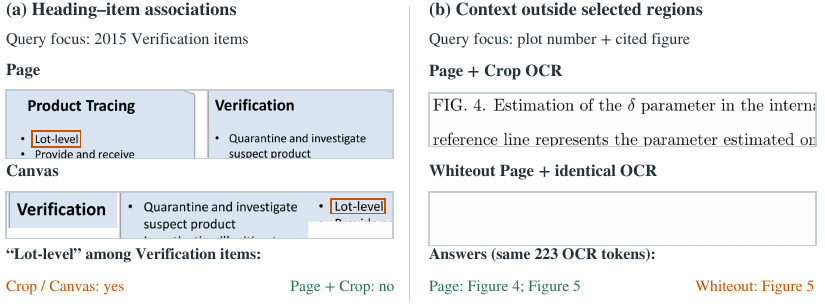}
        \caption{\textbf{The original page keeps context that crops, canvases, and whiteout can lose.}
    \textbf{(a)} ``Lot-level'' belongs to Product Tracing in the page, but appears among Verification items in Crop and Canvas responses.
    Page + Crop excludes it from those items.
    \textbf{(b)} The source caption labeled ``FIG. 4'' is present in Page but removed in Whiteout.
    Their identical OCR does not contain it.
    The displayed answer excerpts concern the requested figure numbers.
    Document excerpts are enlarged for readability, and colored outlines are explanatory annotations.}
    \label{fig:clear-context}
\end{figure}

\paragraph{Heading--item associations.}
The ViDoRe-v3 question in Figure~\ref{fig:clear-context}(a) asks for five Verification items and excluded products.
The original page separates Product Tracing and Verification into columns.
Crop and Canvas responses include Product Tracing items such as ``Lot-level'' among the Verification items, while Page + Crop answers correctly.
Inspection of the complete Canvas input confirms that all five requested items and the items from other columns remain present.
The example illustrates a change in heading--item association rather than deletion of the required text, without isolating layout from changes in image size or grouping.

\paragraph{Surrounding information with identical OCR.}
The SciEGQA question in Figure~\ref{fig:clear-context}(b) asks for the figure number of a plot and the aggregation figure cited in nearby text.
Page + Crop OCR answers Figure 4 and Figure 5 correctly, while Whiteout + the same OCR answers only Figure 5.
The plot and nearby paragraph remain in Whiteout, but the Figure 4 caption falls outside the selected regions and is removed.
The shared OCR also lacks this caption.
Both inputs use 15,730 VT and the same 223 OCR tokens.
This example illustrates information that the original page retains and that the shared selected-region OCR text alone does not replace.

\clearpage
\section[Implementation Details]{Implementation Details}
\label{app:setup}

This section describes the controlled conditions for the analyses in Section~\ref{sec:post_retrieval_analysis}, the answer evaluation protocol shared by all experiments, and the end-to-end RAG configurations.

\subsection[Controlled conditions and evaluation]{Controlled conditions and evaluation}
\label{app:controlled_construction}

\paragraph{Data and page views.}
The SciEGQA comparisons in Section~\ref{sec:post_retrieval_analysis} and Figure~\ref{fig:post_retrieval_grid} use the same 1,623 queries~\citep{yu2026sciegqadatasetscientificevidencegrounded} in every condition, with the retrieved pages fixed to the oracle pages of each query.
ViDoRe-v3~\citep{loison-etal-2026-vidore} contains English and French document datasets and supports monolingual evaluation as well as multilingual evaluation with queries in six languages.
We use the English monolingual setting on all five public English-language datasets, with 1,489 queries in total.
For the Loud and Clear comparisons in Section~\ref{sec:post_retrieval_analysis}, we exclude query ID 239 in the Industrial reports dataset due to missing ground-truth bounding boxes, leaving a base cohort of 1,488 queries.
MinerU~\citep{wang2024mineruopensourcesolutionprecise} parses each retrieved page into paragraph, table, figure, and caption regions.
We retain regions matched to the dataset's ground-truth evidence boxes and crop them directly from the original page image.
When no matched MinerU region exists anywhere in a query, we use raw ground-truth boxes as fallback regions, following Appendix~\ref{app:clear-build}.
All full-page views precede all localized views, such as crops and canvases, in the generator input of each query.

\paragraph{Visual token budgets.}
We measure the visual token budget by the total number of visual tokens produced across all input images by the generator's image processor.
For image-only conditions, the budget is the total number of visual tokens produced across all input images by the generator's image processor. 
For conditions with OCR, we use a shared input budget that counts both OCR text tokens and visual tokens; VT refers specifically
to visual tokens from page images (Appendix~\ref{app:clear-construction}).
The target budget per query is $B\in\{1\mathrm{K},2\mathrm{K},4\mathrm{K},8\mathrm{K},16\mathrm{K},32\mathrm{K},64\mathrm{K}\}$, where $1\mathrm{K}=1{,}000$ tokens.
For the page--crop mixture sweep in Figure~\ref{fig:post_retrieval_grid}(b), the fraction assigned to full-page views is $\alpha\in\{0,.25,.50,.75,1\}$, with $\alpha=1$ and $\alpha=0$ denoting page-only and crop-only inputs.
Full-page views and crops receive target budgets of $\alpha B$ and $(1-\alpha)B$, respectively.
For each query, we apply one continuous resizing factor to each view group and quantize the resulting image sizes to the processor's spatial grid.
We then choose the feasible sizes that minimize the absolute difference between the resulting visual token count and the target budget $B$.
The budget is therefore a target rather than a hard cap, and small overshoots or undershoots are permitted.
Across the 35 combinations of budget and mixture, the maximum absolute deviation of a condition's mean actual token count from its target is 25.72 tokens.
For the 16K comparison in Figure~\ref{fig:post_retrieval_grid}(b), the mean actual counts are 15,993.23 for page-only input and 16,010.19 for the best observed mixture ($\alpha=.25$).

\paragraph{Generation.}
\label{app:qa_generation}
Generation uses Qwen3.6-35B-A3B~\citep{qwen36_35b_a3b} revision \texttt{995ad96eacd98c81e\allowbreak{}d38be0c5b274b040\allowbreak{}31597b0} in BF16 with temperature zero and thinking disabled.
We use vLLM~\citep{10.1145/3600006.3613165} version 0.24.0 with a maximum context length of 262,144 tokens and allow up to 1,024 generated tokens.
The QA instructions and decoding configuration are fixed across all conditions.
OCR conditions additionally prepend the extracted OCR text to the QA prompt.
We use the ViDoRe-v3 generation prompt~\citep{loison-etal-2026-vidore}:
\begin{quote}
\small
\begin{verbatim}
Keep the response concise when appropriate.
Output only the answer.

Question: {question}
Answer:
\end{verbatim}
\end{quote}

\paragraph{Evaluation.}
We score generated answers using the common evaluation protocol in Appendix~\ref{app:answer_evaluation}.
The accuracy reported in the text and Figure~\ref{fig:post_retrieval_grid} is the strict accuracy defined there.

\subsection{Answer evaluation protocol}
\label{app:answer_evaluation}

\paragraph{Judge and inputs.}
The controlled analyses in Section~\ref{sec:post_retrieval_analysis} and the end-to-end comparisons in Section~\ref{sec:rag_evaluation} use the same reference-based judge, \texttt{gpt-5.6-sol}, with medium reasoning effort.
The judge receives the question, reference answer, and predicted answer as text, without document images or method identifiers.
We submit the prompt below as a user message with no additional system instructions, and request structured output with exactly two fields, \texttt{judgment} and \texttt{reason}.
The first field must be \texttt{correct}, \texttt{partial}, or \texttt{incorrect}, and the second contains a brief explanation that names the issue behind a partial or incorrect judgment.

\paragraph{Judgment criteria.}
The judge considers question suitability and reference accuracy equally.
Question suitability requires a clear answer at the scope and specificity requested by the question, while reference accuracy requires semantic agreement with the reference, including the values, entities, relationships, conditions, and additional claims in the answer.
A prediction is \texttt{correct} only when both criteria are fully satisfied.
It is \texttt{partial} when one criterion is fully satisfied and the other is partially satisfied, with a specific reason for the shortfall.
It is \texttt{incorrect} when either criterion is not satisfied at all or neither is fully satisfied.
Equivalent wording and exactly equivalent numerical or unit representations are accepted, but rounding or approximation is accepted only when explicitly allowed by the question or reference answer.

\paragraph{Scores and aggregation.}
For query $i$, let $y_i$ be its final judgment and define the strict score as
\begin{equation}
    z_i = \mathbf{1}\{y_i = \texttt{correct}\},
    \qquad
    \operatorname{Accuracy}(\%) = \frac{100}{N}\sum_{i=1}^{N} z_i,
    \label{eq:strict_accuracy}
\end{equation}
where $N$ is the number of evaluated queries in that benchmark and condition.
Both \texttt{partial} and \texttt{incorrect} therefore receive zero credit.
Each query contributes one final judgment to the reported accuracy.
For ViDoRe-v3, we pool the queries from the five English-language datasets before computing accuracy, giving each query equal weight.
The end-to-end evaluation uses $N=1{,}623$, $1{,}489$, and $1{,}091$ for SciEGQA, ViDoRe-v3, and MMLongBench-Doc, respectively.
The controlled analyses use the query cohorts specified for their respective experiments.
Displayed percentage-point differences are computed by subtracting accuracies rounded to two decimal places.
Bootstrap intervals are computed from unrounded query-level scores.

\paragraph{Judge prompt.}
The following is the full prompt template, with line wrapping adjusted for readability.
We replace \texttt{\{evaluation\_input\}} with a JSON object containing the string fields \texttt{question}, \texttt{reference\_answer}, and \texttt{predicted\_answer}.

\begingroup
\footnotesize
\begin{verbatim}
You are an expert judge evaluating a predicted answer using a question
and a reference answer.

Evaluate the predicted answer using two equally important criteria:

1. Question suitability:
   Whether the predicted answer directly and clearly answers what the
   question asks, at the
   requested scope and level of specificity, and communicates a
   discernible final answer.

2. Reference accuracy:
   Whether the predicted answer is semantically equivalent to the
   reference answer with respect
   to what the question asks. The values, entities, relationships,
   conditions, and any additional
   claims that contribute to the predicted answer must be accurate
   according to the reference
   answer.

The question determines what must be answered, while the reference
answer determines what is
accurate.

Judge semantic meaning rather than wording, structure, format, or
length. Accept semantically
equivalent expressions and exactly equivalent numerical or unit
representations.

Do not accept rounded or approximate values unless the question
explicitly requests rounding,
approximation, a range, or a specific level of precision, or the
reference answer explicitly
provides an approximation or range.

Evaluate only the predicted answer actually provided. Additional
content or apparent truncation
matters only when it affects question suitability or reference
accuracy.

### Evaluation Criteria

- correct:
  The predicted answer fully satisfies both question suitability and
  reference accuracy.

- partial:
  The predicted answer fully satisfies one criterion and partially
  satisfies the other. There must
  be a clear and specific reason why the partially satisfied criterion
  is not fully satisfied.

- incorrect:
  The predicted answer does not satisfy one criterion at all,
  regardless of how well it satisfies
  the other; or neither criterion is fully satisfied.

The reason must briefly identify the specific issue that prevents a
partial or incorrect predicted
answer from being correct.

### Input Data

EVALUATION_INPUT_JSON:
{evaluation_input}

### Output Format

Return only one JSON object with exactly these two fields:

{
  "judgment": "<correct, partial, or incorrect>",
  "reason": "<a very brief explanation>"
}
\end{verbatim}
\endgroup

\subsection{End-to-end RAG configurations}
\label{app:rag_configurations}

Checkpoint links identify the evaluated model revisions for Table~\ref{tab:rag-top4}.

\paragraph{Common generation settings.}
All methods using Qwen3.6-35B-A3B share the checkpoint revision and QA instruction in Appendix~\ref{app:qa_generation}.
Generation uses BF16, temperature zero, disabled thinking, a maximum context length of 65,536 tokens, and an output limit of 3,072 tokens.
The chat template assigns sequential image labels such as \texttt{Picture 1:} in input order.

\paragraph{Page retrieval baselines.}
ColQwen3~\citep{huang2025beyond} uses \href{https://huggingface.co/TomoroAI/tomoro-colqwen3-embed-8b/tree/49658433ec26b7d6ee8225972e2e33577db9f35f}{Tomoro ColQwen3-Embed-8B}, with a retriever limit of 1,280 visual tokens per page.
ColPali~\citep{ICLR2025_99e9e141} uses the \href{https://huggingface.co/vidore/colpali-v1.2/tree/6b89bc63c16809af4d111bfe412e2ac6bc3c9451}{ColPali-v1.2 adapter} with the \href{https://huggingface.co/vidore/colpaligemma-3b-pt-448-base/tree/30ab955d073de4a91dc5a288e8c97226647e3e5a}{ColPaliGemma-3B base} and \texttt{colpali-engine} 0.3.13.
Both baselines rank candidate pages by MaxSim and supply the top four full-page images to the generator in retrieval order.
Their generation processors use bounds of 3,136--1,003,520 pixels per image.

\paragraph{M3DocRAG.}
M3DocRAG~\citep{Cho_2025_ICCV} uses its \href{https://huggingface.co/vidore/colpali-v1.2/tree/6e6cf48566f6fbb8eb429a4bf90f39006d703317}{ColPali-v1.2 adapter revision}, the same base, and \texttt{colpali-engine} 0.3.1.
We follow its PDF processing procedure by rendering all pages at 144 DPI in RGB and resizing them to the document's modal page size before retrieval.
The top four retrieved pages are passed to the common generator in retrieval order, with generation processor bounds of 3,136--12,845,056 pixels for each image.

\paragraph{RegionRAG.}
RegionRAG~\citep{li2026regionrag} uses \href{https://huggingface.co/Aeryn666/RegionRet/tree/79f2d345b30967c1a226812435fc18fba1956828}{RegionRet} with the \href{https://huggingface.co/vidore/colqwen2.5-base/tree/92908120384b7a2110c5beda3ab29cbdb2c08e49}{ColQwen2.5 base}.
We use a saliency threshold of 0.25, a neighborhood radius of one, and maximum saliency for region scoring.
The four highest-scoring regions are selected across the entire candidate corpus, and crops from the same page are grouped following the released implementation.
Each crop is processed within 4,096--262,144 pixels.
Thus, its retrieval count specifies regions rather than pages.

\paragraph{ViSAR.}
ViSAR~\citep{mialland2026visartrainingfreeadaptivekretrieval} uses the same ColPali-v1.2 adapter revision and ColPaliGemma-3B base as the ColPali baseline, with its released image prompt prefix and adaptive selector.
Pages are rendered at 144 DPI without normalizing their sizes within a document.
We use \texttt{top\_T}=50, \texttt{block\_size}=4, \texttt{weight\_thr}=0, $\gamma=100{,}000$, and $\epsilon=0$.
The selector determines the number of pages adaptively, capped at four for this comparison.
Selected pages pass through its 55-million-pixel aggregate guard and then through the common generator's processor with per-image bounds of 3,136--12,845,056 pixels.

\paragraph{VDocRAG.}
VDocRAG~\citep{Tanaka_2025_CVPR} uses \href{https://huggingface.co/microsoft/Phi-3-vision-128k-instruct/tree/ed2772fabe9dc9acd0caad54b62761d92520cc44}{Phi-3-vision-128k-instruct} with separate \href{https://huggingface.co/NTT-hil-insight/VDocRetriever-Phi3-vision/tree/fcf5010505babd789f77d8f008a59f4d72d21805}{VDocRetriever} and \href{https://huggingface.co/NTT-hil-insight/VDocGenerator-Phi3-vision/tree/2fa451b2062b424685bd8a33a6a2d10cfc197cfa}{VDocGenerator} LoRA adapters.
Pages are resized to $1{,}344\times1{,}344$ pixels for its native processor, and the top three retrieved pages are passed to the generator in retrieval order.
We retain the released concise-answer prompt and greedy decoding with a 64-token output limit.

\paragraph{SV-RAG-InternVL2.}
We evaluate the published R5 configuration of SV-RAG~\citep{ICLR2025_0d40c7b4} with the trained \href{https://huggingface.co/puar-playground/Col-InternVL2-4B/tree/0dbd93b1a51d154c3b899e61029b637a9c147752}{Col-InternVL2-4B retrieval adapter} and \href{https://huggingface.co/OpenGVLab/InternVL2-4B/tree/6eb21ec278fbdb03b751a3d48d8dfc9724e05608}{InternVL2-4B} for generation without additional QA fine-tuning.
The top five pages are selected, then supplied in corpus order following the released implementation.
Each page is divided into up to twenty $448\times448$ tiles, with a thumbnail added when more than one tile is used.
We retain the released QA prompt and use greedy decoding, with an output limit of 1,024 generated tokens per answer.

\paragraph{EVisRAG-7B.}
EVisRAG-7B~\citep{sun2026visrag20mitigatingvisualhallucinations} uses \href{https://huggingface.co/openbmb/VisRAG-Ret/tree/95ef596df871b606167cb7e4b7215caf1bfdf761}{VisRAG-Ret} and the \href{https://huggingface.co/openbmb/EVisRAG-7B/tree/96159ef32143770d61167fefd514b99117dc7747}{EVisRAG-7B checkpoint} released on October 2, 2025, based on Qwen2.5-VL-7B.
The top three pages are provided in retrieval order, with processor bounds of 3,136--3,920,000 pixels per image.
We retain the released evidence-guided prompt, place its text before the images, and use temperature zero, repetition penalty 1.05, and a 2,048-token output limit.
We extract text after the first \texttt{<answer>} tag, stopping at \texttt{</answer>} or the response end.
Responses without an opening tag are retained in full.

\paragraph{Additional generators.}
For Table~\ref{tab:rag-generators}, we use Qwen2.5-VL-7B-Instruct and Qwen2.5-VL-32B-Instruct with their native processors, tokenizers, and chat templates.
Both use 14-pixel patches merged $2\times2$, giving one image VT per $28\times28$ pixels.
ColQwen3 supplies the same top-four pages in retrieval order, smart-resized with per-image bounds of 3,136--1,003,520 pixels.
For LoC, each baseline dimension is multiplied by $s=2.25$ and rounded to the nearest multiple of 28 using Python's \texttt{round}; the resulting image VT sum defines the query-specific budget $B$.
OCR text, including its evidence labels, is tokenized first; pages receive the remaining budget through common-scale resizing with dimensions floored to the 28-pixel grid, leaving quantization slack unused.
Thus, image VT plus OCR tokens is at most $B$; question, instruction, chat wrappers, and output tokens are excluded.
Images precede OCR and the unchanged QA instruction, with native \texttt{Picture 1:} labels and the system message \texttt{You are a helpful assistant.}
Both models use BF16, vLLM 0.24.0, TP=1, greedy decoding (temperature 0, top-$p$ 1, seed 0), \texttt{generation\_config=vllm}, a 65,536-token context, and a 3,072-token output limit.

\clearpage
\section[Additional Results on End-to-End RAG Systems]{Additional Results on End-to-End RAG Systems}
\label{app:end-to-end-analysis}

\subsection{Paired comparisons and uncertainty}
\label{app:end-to-end-paired}

The three conditions in Table~\ref{tab:rag-top4} cover identical query sets, questions, and reference answers within each benchmark.
Recomputing accuracy from their saved predictions and final judgments reproduces all nine reported accuracies.
No query intersection filtering is needed, and answers reaching the output limit remain included.
Unlike the controlled comparisons in Section~\ref{sec:post_retrieval_analysis}, the end-to-end evaluation includes all 1,489 ViDoRe-v3 queries because it does not require ground-truth bounding boxes.

We compare ColQwen3 with \sname{} w/o Clear (Loud), \sname{} w/o Clear with \sname{} (Clear), and ColQwen3 with \sname{} (Total).
Only \texttt{correct} receives credit; \texttt{partial} and \texttt{incorrect} both receive zero.
For each comparison, we estimate a paired percentile bootstrap interval by resampling queries with both conditions' outcomes together.
All bootstrap analyses in this appendix use 20,000 resamples, NumPy 2.3.5 with linear percentile interpolation, and fixed comparison-specific seeds derived from the base seed 20260925.
Intervals are computed from unrounded query scores; displayed accuracy changes follow the rounding convention in Appendix~\ref{app:answer_evaluation}.
These marginal intervals are not corrected for multiple comparisons and describe query-sampling uncertainty conditional on the saved answers and judgments, rather than variation across independent model runs.

\begin{table}[t]
    \centering
    \setlength{\belowcaptionskip}{\abovecaptionskip}
    \setlength{\abovecaptionskip}{0pt}
    \caption{\textbf{Paired comparisons quantify the end-to-end gains.}
    Improved and Regressed count non-correct-to-correct and correct-to-non-correct changes, respectively.
    Changes and 95\% confidence intervals are in percentage points.
    Loud, Clear, and Total denote the comparisons defined above.}
    \label{tab:loc-paired-saved}
    \begingroup
    \small
    \setlength{\tabcolsep}{4pt}
    \resizebox{\linewidth}{!}{%
    \begin{tabular}{@{}llrrrrr@{}}
        \toprule
        Dataset & Comparison & $N$ & Improved & Regressed & $\Delta$ & 95\% CI \\
        \midrule
        SciEGQA & Loud  & 1623 & 176 & 74 & +6.29 & $[4.37, 8.19]$ \\
                & Clear & 1623 & 114 & 72 & +2.59 & $[0.99, 4.25]$ \\
                & Total & 1623 & 223 & 79 & +8.88 & $[6.84, 10.91]$ \\
        \midrule
        ViDoRe-v3 & Loud  & 1489 & 103 & 65 & +2.55 & $[0.87, 4.30]$ \\
                  & Clear & 1489 & 93 & 85 & +0.54 & $[-1.21, 2.28]$ \\
                  & Total & 1489 & 114 & 68 & +3.09 & $[1.34, 4.84]$ \\
        \midrule
        MMLongBench-Doc & Loud  & 1091 & 57 & 36 & +1.93 & $[0.18, 3.67]$ \\
                       & Clear & 1091 & 71 & 46 & +2.29 & $[0.37, 4.22]$ \\
                       & Total & 1091 & 84 & 38 & +4.22 & $[2.29, 6.14]$ \\
        \bottomrule
    \end{tabular}%
    }
    \endgroup
\end{table}

The paired query-bootstrap intervals for \sname{} over ColQwen3 exclude zero on all three benchmarks (Table~\ref{tab:loc-paired-saved}).
For Clear, the intervals exclude zero on SciEGQA and MMLongBench-Doc, while the smaller gain on ViDoRe-v3 remains uncertain.

\subsection{ViDoRe-v3 domain results}
\label{app:end-to-end-domains}

Clear's contribution varies across the five English-language domains (Table~\ref{tab:loc-vidore-domains-saved}).
Its net gain of eight correct answers combines gains in computer science ($+12$), pharmaceuticals ($+9$), and industrial reports ($+5$) with losses in HR ($-12$) and finance ($-6$).
The full \sname{} method has higher observed accuracy than ColQwen3 in all five domains, with computer science, finance, and pharmaceuticals accounting for 44 of the 46 additional correct answers.
Domain intervals are exploratory, use paired query resampling, and are not corrected for multiple comparisons.

\begin{table}[t]
    \centering
    \setlength{\belowcaptionskip}{\abovecaptionskip}
    \setlength{\abovecaptionskip}{0pt}
    \caption{\textbf{Domain results explain the small pooled Clear gain on ViDoRe-v3.}
    Accuracy is reported in percent; Clear changes and intervals are in percentage points.
    The pooled row weights each query equally.
    Displayed changes subtract rounded accuracies, while intervals use unrounded query scores.}
    \label{tab:loc-vidore-domains-saved}
    \begingroup
    \small
    \setlength{\tabcolsep}{4pt}
    \resizebox{\linewidth}{!}{%
    \begin{tabular}{@{}lrrrrrr@{}}
        \toprule
        Domain & $N$ & ColQwen3 & \sname{} w/o Clear & \sname{} & Clear $\Delta$ & Clear 95\% CI \\
        \midrule
        Computer science & 215 & 59.07 & 59.07 & 64.65 & +5.58 & $[0.47, 10.70]$ \\
        Finance & 309 & 43.69 & 51.13 & 49.19 & -1.94 & $[-5.83, 1.94]$ \\
        HR & 318 & 44.03 & 48.11 & 44.34 & -3.77 & $[-7.86, 0.31]$ \\
        Industrial & 283 & 49.47 & 48.06 & 49.82 & +1.76 & $[-1.77, 5.30]$ \\
        Pharmaceuticals & 364 & 41.76 & 43.41 & 45.88 & +2.47 & $[-0.82, 5.77]$ \\
        \midrule
        Pooled micro & 1489 & 46.61 & 49.16 & 49.70 & +0.54 & $[-1.21, 2.28]$ \\
        \bottomrule
    \end{tabular}%
    }
    \endgroup
\end{table}

\subsection{Document-cluster sensitivity}
\label{app:end-to-end-clusters}

We additionally resample whole documents with replacement, retaining all their query pairs and recomputing query-weighted accuracy differences.
On SciEGQA (80 documents), the 95\% intervals for Loud, Clear, and Total are $[4.28, 8.34]$, $[0.98, 4.15]$, and $[6.55, 11.31]$ percentage points, respectively.
On MMLongBench-Doc (135 documents), they are $[0.00, 3.88]$, $[0.18, 4.47]$, and $[2.33, 6.19]$.
Thus, the Clear and Total intervals remain above zero on both benchmarks, while the lower bound for Loud on MMLongBench-Doc is approximately zero.
We omit document-cluster resampling for ViDoRe-v3 because 130 queries reference multiple documents, so a unique non-overlapping document assignment would require additional assumptions.

\clearpage
\section[Generation-stage Latency across Visual Token Budgets]{Generation-stage Latency across Visual Token Budgets}
\label{app:latency}

The saved SciEGQA runs underlying Figure~\ref{fig:rag-budget-sweep} allow us to examine how generation-stage latency varies with the visual token budget.
We report \sname{} and \sname{} w/o Clear at nine scales, using the same 1,623 queries in every condition (29,214 query records).
Accuracy is computed from the judged predictions of the corresponding timed runs.
All conditions have complete timing records and no final generation failures or timeouts; slow queries and answers reaching the output limit remain included.

\paragraph{Measurement scope.}
The reported latency includes input preprocessing and engine request residence time under concurrent execution.
Preprocessing covers image loading, conversion, resizing, and processor/tokenizer operations.
Engine residence is measured from enqueueing a request to receiving its completed output, including engine queueing and shared execution.
The sum excludes retrieval, region selection, OCR, model loading, producer ready-queue waiting, and some preparation and I/O operations.
It therefore measures neither end-to-end RAG latency nor isolated GPU execution time.
Tables~\ref{tab:loc-latency} and~\ref{tab:loc-latency-reused} report the median and descriptive 95th percentile (P95), computed using linear interpolation.

\paragraph{Execution and comparability.}
The runs share the recorded GPU model and generation settings, but their workloads differ.
Each worker uses one NVIDIA RTX PRO 6000 Blackwell Max-Q Workstation Edition GPU with Qwen3.6-35B-A3B, vLLM 0.24.0, BF16, and tensor parallelism of one.
Generation settings follow Appendix~\ref{app:rag_configurations}; continuous batching permits at most 128 sequences and 16,384 batched tokens, rather than a fixed batch size.
Table~\ref{tab:loc-latency} reports the SciEGQA-only sweeps.
Table~\ref{tab:loc-latency-reused} reports the two reused $s=2.25$ sessions, each of which processed 4,203 queries across benchmarks in a separate run, together with the reused $s=1.0$ page baseline.
The reused $s=1.0$ page baseline also comes from a mixed-benchmark session and differs in input preparation and the absence of the explicit answer warm-up used by the other runs.
Model loading precedes query timing, and no benchmark queries are excluded as warm-up.
Host identity and instantaneous resource load were not recorded, so these are descriptive comparisons of archived runs, not controlled speedup measurements.

\begin{table}[t]
    \centering
    \setlength{\belowcaptionskip}{\abovecaptionskip}
    \setlength{\abovecaptionskip}{0pt}
    \caption{\textbf{Median generation-stage latency increases with the visual token budget in both SciEGQA-only sweeps.}
    Every row includes all 1,623 queries.
    Mean used budget counts image visual tokens plus OCR text tokens, excluding the question, instructions, and chat delimiters. K denotes 1,000 tokens.
    Unlike the target cap used in Figure~\ref{fig:rag-budget-sweep}, this column reports actual usage, which can be slightly lower for \sname{} due to token quantization.
    Mean output length is included because decoding workload also varies.
    Reused session records are reported separately in Table~\ref{tab:loc-latency-reused}.}
    \label{tab:loc-latency}
    \small
    \setlength{\tabcolsep}{3.5pt}
    \renewcommand{\arraystretch}{1.06}
    \begin{tabular}{@{}lrrrrrr@{}}
        \toprule
        Variant & $s$ & \shortstack{Mean used\\budget (K)} & \shortstack{Mean output\\tokens} & \shortstack{Median\\(s)} & \shortstack{P95\\(s)} & \shortstack{Acc.\\(\%)} \\
        \midrule
        \multicolumn{7}{@{}l}{\textit{SciEGQA-only sweep sessions}} \\
        \sname{} w/o Clear & 1.25 & 5.959 & 69.37 & 34.02 & 191.71 & 72.15 \\
        \sname{} w/o Clear & 1.50 & 8.429 & 56.75 & 79.59 & 241.53 & 73.81 \\
        \sname{} w/o Clear & 1.75 & 11.594 & 49.83 & 121.65 & 324.95 & 75.11 \\
        \sname{} w/o Clear & 2.00 & 15.190 & 53.78 & 205.42 & 363.99 & 75.23 \\
        \sname{} w/o Clear & 2.50 & 23.927 & 39.09 & 416.45 & 503.76 & 75.11 \\
        \sname{} w/o Clear & 2.75 & 28.662 & 32.42 & 572.89 & 646.46 & 75.17 \\
        \sname{} w/o Clear & 3.00 & 34.177 & 33.42 & 745.66 & 792.53 & 75.23 \\
        \sname{} & 1.00 & 3.762 & 82.36 & 16.44 & 41.29 & 65.87 \\
        \sname{} & 1.25 & 5.901 & 84.43 & 17.83 & 137.58 & 72.70 \\
        \sname{} & 1.50 & 8.354 & 63.23 & 56.00 & 237.05 & 75.05 \\
        \sname{} & 1.75 & 11.498 & 58.87 & 96.75 & 298.27 & 75.97 \\
        \sname{} & 2.00 & 15.079 & 70.98 & 159.90 & 418.36 & 76.65 \\
        \sname{} & 2.50 & 23.776 & 68.18 & 343.44 & 666.16 & 77.88 \\
        \sname{} & 2.75 & 28.500 & 70.27 & 482.30 & 729.37 & 78.93 \\
        \sname{} & 3.00 & 33.998 & 70.99 & 658.35 & 1000.64 & 78.50 \\
        \bottomrule
    \end{tabular}
\end{table}

\begin{table}[t]
    \centering
    \setlength{\belowcaptionskip}{\abovecaptionskip}
    \setlength{\abovecaptionskip}{0pt}
    \caption{\textbf{Execution differences limit direct latency comparisons across reused sessions.}
    Every row reports all 1,623 SciEGQA queries from its session.
    Column definitions follow Table~\ref{tab:loc-latency}.
    The two $s=2.25$ variants were run in separate mixed-benchmark sessions.
    The $s=1.0$ page baseline also comes from a mixed-benchmark session, with different input preparation and no explicit answer warm-up.
    These records do not establish a controlled latency ratio between the page baseline and \sname{}.}
    \label{tab:loc-latency-reused}
    \small
    \setlength{\tabcolsep}{3.5pt}
    \renewcommand{\arraystretch}{1.06}
    \begin{tabular}{@{}lrrrrrr@{}}
        \toprule
        Variant & $s$ & \shortstack{Mean used\\budget (K)} & \shortstack{Mean output\\tokens} & \shortstack{Median\\(s)} & \shortstack{P95\\(s)} & \shortstack{Acc.\\(\%)} \\
        \midrule
        \multicolumn{7}{@{}l}{\textit{Reused $s=2.25$ sessions (separate run per variant)}} \\
        \sname{} w/o Clear & 2.25 & 19.271 & 44.28 & 264.91 & 378.91 & 75.48 \\
        \sname{} & 2.25 & 19.137 & 63.23 & 246.34 & 467.44 & 78.07 \\
        \midrule
        \multicolumn{7}{@{}l}{\textit{Reused page baseline (different preparation and warm-up)}} \\
        \sname{} w/o Clear & 1.00 & 3.797 & 74.44 & 6.81 & 25.90 & 69.19 \\
        \bottomrule
    \end{tabular}
\end{table}

\paragraph{Observed tradeoff.}
Within each dedicated sweep in Table~\ref{tab:loc-latency}, median latency increases monotonically with the visual token budget, while accuracy gains diminish at larger budgets.
For \sname{} w/o Clear, increasing $s$ from 1.75 to 3.0 changes accuracy from 75.11\% to 75.23\%, while median latency rises from 121.65 to 745.66 seconds.
For \sname{}, increasing $s$ from 2.75 to 3.0 changes accuracy from 78.93\% to 78.50\%, while median latency rises from 482.30 to 658.35 seconds.
Across the seven scales shared by the dedicated sweeps, \sname{} has lower median latency, but its P95 is higher at $s=2.0$, 2.5, 2.75, and 3.0.
For example, at $s=3.0$, its median is 658.35 versus 745.66 seconds, whereas its P95 is 1,000.64 versus 792.53 seconds.
Thus, the results show an observed accuracy--latency tradeoff without establishing a uniform latency advantage for \sname{} or a latency-optimal choice of $s$.
Input modality, output length, batching, and queueing also vary across runs; the latency differences cannot be attributed solely to the visual token budget.
End-to-end RAG latency and peak memory remain unmeasured.

\end{document}